\documentclass[3p,times,procedia]{elsarticle}
\usepackage{ecrc}

\volume{00}

\firstpage{1}

\journalname{Transportation Research Procedia}

\runauth{Om Chiddarwar}

\jid{trpro}

\usepackage{amsthm}
\usepackage{siunitx}
\usepackage[labelfont=bf, textfont=bf]{caption} 
\usepackage{subcaption}
\usepackage{times}
\usepackage{geometry}
\usepackage{parskip}
\usepackage{float}  
\usepackage{comment}
\usepackage{indentfirst}  
\usepackage{graphicx} 
\usepackage{amsmath, amssymb, amsthm}
\usepackage{siunitx}
\usepackage{graphicx}
\usepackage{float}
\usepackage[labelfont=bf, textfont=bf]{caption}
\usepackage{subcaption}
\usepackage{geometry}
\usepackage{parskip}
\usepackage[figuresright]{rotating}
\usepackage[bookmarks=false]{hyperref}
\hypersetup{colorlinks, linkcolor=blue, citecolor=blue, urlcolor=blue}

\biboptions{authoryear}

\usepackage[figuresright]{rotating}

\makeatletter
\g@addto@macro\normalsize{%
\setlength\abovedisplayskip{6pt}
\setlength\belowdisplayskip{6pt}
\setlength\abovedisplayshortskip{4pt}
\setlength\belowdisplayshortskip{4pt}
}
\makeatother
\usepackage[bookmarks=false]{hyperref}
\hypersetup{colorlinks,
linkcolor=blue,
citecolor=blue,
urlcolor=blue}

\begin{document}
\begin{frontmatter}

	\dochead{World Conference of Transport Research, Toulouse 2026 (WCTR 2026 Toulouse)}%
	
	\title{Predicting Spatiotemporal Mobile Sensing-Based PM\textsubscript{2.5} Concentrations Using Low-Rank Adapted Spatially Attentive Graph Neural Network}



	\author[a]{Om Chiddarwar} 
	
	\author[b]{Priyanka Mandal}
	\author[b]{Praveen Kumar Chandaliya}
	\author[b]{Shriniwas Arkatkar\corref{cor1}}
	
	\address[a]{IIITDM Kurnool, Kurnool, Andhra Pradesh 518002, India}
	\address[b]{Sardar Vallabhbhai National Institute of Technology Surat, Gujarat 394210, India}

	\begin{abstract}
		Urban air quality can vary significantly along transit corridors, necessitating high-resolution monitoring. This work introduces a novel mobile-sensing dataset from Surat, Gujarat, India, comprising PM\textsubscript{2.5} concentrations, meteorological variables (temperature, humidity, wind speed, wind direction), and land-use features. To represent the spatiotemporal data as a graph, two node-definition strategies were used: (i) \textbf{uniform segmentation} (200–400~m intervals) and (ii) \textbf{DBSCAN clustering} to adaptively group dense observations. For each node, rolling mean and standard deviation of meteorological variables were computed. To model this high-dimensional data, we propose a \textbf{SA-GNN} for fine-grained, short-term PM\textsubscript{2.5} forecasting and hotspot identification. We compared SA-GNN with LSTM, RNN, GRU, and ANN models. These models worked well on low-detail data but had trouble handling the fast-changing patterns in urban air quality. SA-GNN employs \textbf{cluster-specific GRUs} to capture localized temporal dependencies and a \textbf{Graph Attention Network} to learn spatial heterogeneity. This hybrid architecture effectively models rapid fluctuations and complex spatial interactions. On our dataset SA-GNN achieved \textbf{R² = 0.95, RMSE = 6.8, MAE = 4.2~\si{\micro\gram\per\meter\cubed} } outperforming all baselines. Combining spatial clustering with adaptive attention significantly improves forecasting, enabling real-time, fine-grained monitoring and supporting personalized exposure tracking and timely alerts for healthier cities.
		
	\end{abstract}
	
	\vspace{-2em}
	\cortext[cor1]{}
\end{frontmatter}


\email{523cs0001@iiitk.ac.in}



\section{Introduction}\label{sec:introduction}
\hspace{2em} Air pollution is still among the greatest public and environmental health issues of the 21st century and is projected to account for millions of deaths and disease states like respiratory infections, asthma and cardiovascular diseases every year. Having the capability to penetrate the lungs or even into the blood, fine particulate matter (PM$_{2.5}$), made up of (particles with a diameter of 2.5 micrometers or less), is especially deadly among air emissions~\cite{pope2006health, cohen2017estimates}. Seven million fatalities annually due to PM$_{2.5}$ exposure are estimated by the World Health Organization (WHO).

\hspace{2em} In rapidly motorizing regions, vehicular emissions along urban transport corridors have emerged as a dominant source of PM$_{2.5}$, disproportionately affecting commuters and pedestrians. Air pollution is at extreme levels in South and Southeast Asia, driven by economic development, urbanization, intense traffic congestions, and seasonal biomass burning. Delhi and Kanpur have continuously recorded PM$_{2.5}$ at much higher rates than the global safe level in India, typically between 60 and 300 µg/m$^3$, compared to the World Health Organization standard for an average of 15 µg/m$^3$ for 24 hours~\cite{balakrishnan2019impact, shaddick2018data, hammer2020global}. While northern hotspots are better documented, fast-expanding industrial cities like Surat in Gujarat are increasingly vulnerable. As a humid coastal urban agglomeration with high textile and manufacturing density and rapidly rising vehicular emissions, Surat has elevated levels of PM$_{2.5}$ particularly during winter, when air stagnates, severely limiting the pollutant dispersion. Real-time accurate forecasting of PM$_{2.5}$ at high spatial and temporal resolutions is therefore vital, not only for regulatory measures and public health, but also for enabling transport system resilience, commuter exposure minimization, and informed mobility decisions. Fine-scale predictions enable city governments to provide location-based pollution alerts and take location-specific measures, besides helping individuals make educated decisions regarding outdoor activities. Additionally, such forecasts are critical for personal use, such as route-based air quality alerts for pedestrians, cyclists, motorized two wheeler and public transit users on prominent city corridors in Delhi, Mumbai, Kolkata, and Chennai.

\hspace{2em} Despite growing interest in fine-scale air quality prediction in transport planning and traffic-related air quality management, significant challenges persist due to the complex, dynamic nature of urban traffic corridor pollution level.  Concentrations of pollutants may change drastically over extremely short distances and timescales, influenced by a combination of local emissions, meteorological variability, and geography~\cite{liang2015assessing}. Static monitoring stations employed by urban governments lack good spatial resolution and fail to capture micro-level fluctuations of interest to individuals~\cite{kumar2015rise, apte2017high}.  particularly along high-traffic arterials where exposure peaks occur near intersecions and idling zones. Although mobile sensing is a more micro-level view, it introduces other issues such as non-uniform sampling rates, GPS noise, and missing values~\cite{castell2017can}. Besides that, the variability of PM$_{2.5}$ values also arises from the non-homogeneity of exogenous conditions such as climatic, atmospheric, and weathering conditions and transport-related factors such as traffic density and land use patterns~\cite{zhou2020forecasting}. Nonetheless, the dynamic and non-linear nature of the relationship between PM$_{2.5}$ and its determinate factors makes it difficult to predict. Traditional statistical models, while computationally fast, fail to capture the spatiotemporal interactions driven by traffic flow and urban mobility patterns. The traditional models such as Autoregressive Integrated Moving Average (ARIMA), Kalman Filters, and Multiple Linear Regression (MLR) have previously been extensively utilized in forecasting air quality since they are computationally inexpensive and simple to interpret~\cite{yi2018review}.

\hspace{2em} To overcome such limitations, machine learning techniques such as Random Forests, Support Vector Regression, and Gradient Boosting Machines have been explored~\cite{zheng2013u, gao2020predicting}. Such a model can learn more intricate relationships, but it does not necessarily perform as well with sequential dependencies or spatial generalizability. Deep learning models such as CNNs, RNNs, and Transformers are observed to learn intricate temporal structures but generally fail to capture spatial heterogeneity, particularly for non-uniform urban environments~\cite{zhou2024deep, cui2023comparative, samal2024autotcn}. Combined and fusion-based deep learning models through a combination of convolutional layers with recurrent (e.g., ConvLSTM) or attention-based modules (e.g., CNN-GRU) have been put forward to better model the spatiotemporal dependencies~\cite{ma2020learning, chen2022hybrid}. These methods are still constrained in dealing with mobile node topologies, non-uniform sampling rates in space, and noise structures of mobile sensing environments. Their effectiveness worsens when extended beyond the initial static station-based or grid-based settings in which they were originally developed.

\hspace{2em} Most of today's GNN models are founded on fixed monitoring stations and are not designed to deal with mobile sensing scenarios involving changing spatial patterns. To bridge this gap, a novel Spatially Attentive Cluster-based Graph Neural Network (SA-GNN) is proposed, specifically tailored for fine-grained PM$_{2.5}$ prediction based on mobile sensors. The approach constructs a dynamic pseudo-station graph from two complementary methods: fixed-length trajectory segmentation and DBSCAN-based spatially dense measurement clustering. For each resulting node, temporal relations are remembered by Gated Recurrent Units (GRUs) to ensure resilience to missing or corrupted data. Spatial correlations are stored in a \textbf{GAT-based mechanism} dynamically weighing local clusters, enabling the model to learn local environmental interaction and non-uniform pollutant dispersion. The approach was implemented with a home-brewed, bike-mounted mobile sensing platform, collecting over 14 km of air quality and meteorological sensor data in Surat, India. Mobile sensing has a number of advantages over fixed monitoring stations, including enhanced spatial resolution and the ability to offer individualized exposure data~\cite{snyder2013changing}. Feature engineering techniques such as rolling statistics and log transformation were employed to regularize the training process and enhance the overall generalizability of the models. Experimental results confirm that the constructed SA-GNN achieves state-of-the-art forecasting performance with an \textbf{$R^2$ = 0.95}, \textbf{RMSE = 6.84~$\mu\text{g/m}^3$}, and \textbf{MAE = 4.19~$\mu\text{g/m}^3$}, outperforming conventional models such as LSTM, GARNN~\cite{wang2020pm25gnn}, and physics-informed GNNs~\cite{wang2023forecasting}. 

In summary, the key contributions of this paper are as follows:
\begin{itemize}
	\item The challenge of real-time fine-grained real-time PM$_{2.5}$ forecasting in dynamic, traffic-induced mobile sensing environments is addressed, where spatial structures and data quality vary over time with commuter movement and congestion patterns.
	\item A novel \textbf{Spatially Attentive Cluster-based Graph Neural Network (SA-GNN)} is proposed, combining dynamic spatial clustering (fixed segmentation and DBSCAN) with attention-based graph learning to model complex spatiotemporal pollution patterns.
	\item The framework leverages \textbf{GRU modules} for robust temporal modeling under irregular sampling and noisy data conditions.
	\item The SA-GNN model is validated on a custom mobile sensing dataset collected over a 14 km urban arterial transport corridor in Surat, India, demonstrating superior performance (\textbf{$R^2$=0.95}, \textbf{RMSE=6.84~$\mu\text{g/m}^3$}, \textbf{MAE=4.19~$\mu\text{g/m}^3$}) compared to LSTM, RNN, and physics-informed GNN baselines.
	\item This work highlights the potential of integrating mobile sensing with advanced graph neural networks for scalable, personalized air quality forecasting and exposure aware mobility planning in urban environments.
\end{itemize}

\vspace{-1.5em}
The paper is structured as follows: Section~\ref{sec:relatedwork} reviews prior research on spatiotemporal pollution forecasting. Section~\ref{sec:problem} outlines the proposed methodology, and Section~\ref{sec:experimental} describes the dataset and experimental setup. Section~\ref{sec:architecture} details the network architecture of the proposed model. Sections~\ref{sec:result1} and~\ref{sec:result2} present the qualitative and quantitative analyses, respectively. Finally, Section~\ref{sec:conclusion} summarizes the key findings and suggests future research directions.

\section{Related Work}\label{sec:relatedwork}
\indent Forecasting  PM2.5 along the urban transport corridors has evolved from classical statistical time series model to advanced deep learning and, more recently, graph neural networks. The motivation behind this has predominantly been the capture of highly nonlinear, spatially correlated, and temporally dynamic pollutant behavior of substances like PM$_{2.5}$.

\subsection{Statistical Forecasting Models}

\indent Early attempts at air quality forecasting were based on traditional time series models like Autoregressive Integrated Moving Average (ARIMA) and Kalman filters~\cite{kumar2011arima, wang2014kalman}. They were good at extracting linear temporal patterns from air pollution data and were comparatively easier to implement. Nevertheless, their stationarity and linearity assumptions restricted them from capturing the complex relationships between PM$_{2.5}$ and emission sources, atmospheric conditions, and city geography. Furthermore, these models were unable to perform multivariate forecasting and could not generalize across varying environmental and climatic conditions
particularly failing to model 
the rapid fluctuations observed along congested urban arterials.

\subsection{Machine Learning Methods: Traditional}\label{subsec:ml}
To address the nonlinearity, traditional ML models incorporated traffic and land use features but 
remained limited in sequential and spatial modeling.
In response to the shortcomings of linear modelling, machine learning methods like Support Vector Machines (SVMs), Random Forests (RF), and Gradient Boosting Machines (GBMs) were embraced~\cite{zheng2013u, gao2020predicting}. These methods brought non-linearity to the modeling procedure and enhanced short-term forecast accuracy. But they were largely based on hand-crafted feature engineering and domain-specific pre-processing. For the purpose of PM$_{2.5}$ prediction, temperature, relative humidity, wind speed and direction, traffic volume, vehicle speed, land use, and day-of-week effects are critical in determining pollutant dynamics. Those variables need to be well-preprocessed using domain expertise by operations like rolling averages, lag features, and categorical encodings in order to identify trends and interactions over time~\cite{yi2018review}. Even with these improvements, such models lack an internal mechanism to model temporal memory and cannot properly learn long-range dependencies or sequence patterns found in air quality measurements. This disadvantage is most serious for the forecasting tasks for non-uniform and asynchronous data collection, as in mobile sensing platforms. In addition, spatial heterogeneity and dynamic correlations among monitoring sites particularly when they use pseudo-nodes based on mobile sensors along transport routes cannot be suitably represented by traditional ML architectures.

\subsection{Deep Learning-Based Methods}\label{subsec:deeplearn}
The advent of deep learning introduced substantial improvements in modeling time-varying pollution dynamics. Long Short-Term Memory (LSTM) and Gated Recurrent Unit (GRU) architectures of Recurrent Neural Networks (RNNs) have been used to model sequential measurements of pollutants and performed better than static models~\cite{li2017lstm}. RNN-based models, though successful, typically suffer from vanishing gradients on long sequences and are unable to take advantage of spatial relationships between air quality monitoring stations.
To mitigate spatial dependencies, spatiotemporal deep learning hybrid models like Convolutional LSTM (ConvLSTM) and 3D-CNNs were introduced~\cite{wen2019deep}.
These models utilize grid-based representations to model temporal and spatial patterns simultaneously. PlumeNet~\cite{alleon2020plumenet}, for instance, uses convolutional layers accompanied by recurrent units to provide fine-grained prediction. Others, like attention-augmented CNN architectures~\cite{zhang2022attentioncnn}, integrate spatial attention mechanisms to capture local air quality variations more efficiently. Yet, these approaches usually assume a uniform grid structures or fixed placements of sensors, and there is a shortfall in dealing with dynamic spatial relationships especially those found in mobile or distributed sensor networks. The most significant limitation of these approaches is the assumption of a stationary graph topology or fixed node positions that fail to generalize to mobile or irregular deployments of sensors.As opposed to the dataset employed in this work, which is built from a mobile urban sensor platform with varying sensing locations across time, the dynamic spatial configuration calls for more adaptive models with the ability to cope with changing graph structures, hence the application of graph neural networks with spatial attention and low-rank adaptation.

\vspace{-1em}
\subsection{Graph Neural Network Based Models}\label{subsec:gnn}
Graph Neural Networks (GNNs) offer a natural framework for modeling irregular spatial structures, 
but most are designed for fixed stations and struggle with mobility-induced topology changes.
The spatiotemporal forecasting of PM$_{2.5}$ patterns is an n-dimensional phenomenon.
Influential factors on its variation encompass exogenous factors like climate and atmospheric environment, and anthropogenic factors like land use, traffic conditions, and presence of intersections. These features add to the complexity and necessitate solid modeling frameworks. In this regard, graph-based models offer an organic vehicle to deal with irregular spatial structures. Early methods such as Diffusion Convolutional Recurrent Neural Networks (DCRNN) and Graph Convolutional LSTM (GC‑LSTM) integrate graph convolution with temporal modeling over fixed station networks~\cite{li2017diffusion, qi2019gclstm}. Later works have integrated domain knowledge and physical principles.PM2.5‑GNN incorporates meteorological features, land-uses, emissions, and altitude as node and edge features, hence improving interpretability ~\cite{wang2020pm25gnn}. Physics-aware models like DPGN and Differentiable‑Physics‑aware Graph Networks embed partial differential equation–based diffusion-advection dynamics within graph models, providing enhanced long-distance forecasting and physical plausibility~\cite{seo2019differentiable, hettige2024airphynet}. AirPhyNet follows this trend with direct modeling of pollutant transport for enhanced accuracy in low-data scenarios~\cite{hettige2024airphynet}, with GraPhy also finding additional theoretical support through chemical process modeling~\cite{theoryguided2021}. Contemporary GNNs resort more and more to attention mechanisms to enhance learning. Transformer-like spatial–temporal models like GT‑GNN and GLSTM‑MA attain enhanced long-horizon predictive performance, albeit assuming fixed station arrangements remain. The increasing use of mobile, wearable, and distributed sensing has created new issues. Fixed station topologies-oriented traditional GNNs perform poorly in dynamic urban deployments. Mobile-centric solutions like Deep‑MAPS and light-weight regression models based on Bayes usually lose spatial structure and have difficulty with irregular sampling or GPS drift ~\cite{yoursource11}.
To solve these challenges, an adaptive  framework, referred to as \textbf{SA-GNN} (Spatially Attentive Cluster-based Graph Neural Network), is proposed.

The framework is combined using dynamic clustering, spatial graph learning, and temporal attention. DBSCAN is utilized to dynamically create clusters from trajectories of mobile sensors, allowing localized and personalized prediction. An irregular sequence handling noisy input and a GRU-based temporal module deal with, while a spatially adaptive learning of influence by a GAT (Graph Attention Network) module is achieved with local station and meteorological data. Prediction resilience is additionally enhanced by a linear attention mechanism tuned through Low-Rank Adaptation (LoRA)~\cite{hu2022lora}. This architecture facilitates high-resolution PM$_{2.5}$ forecasting in mobile and personal sensing environments beyond the capability of current city-scale, static GNN models. 

\vspace{-1.5em}
\section{Problem Formulation}\label{sec:problem}
\noindent In this study, we formulate the fine-grained PM\textsubscript{2.5} forecasting task as a spatiotemporal learning problem on a graph structure. Unlike traditional approaches that rely solely on fixed-location Air Quality Monitoring Stations (AQMS), we treat each spatial segment or dynamically-formed cluster of mobile sensor observation  as a node in the graph.  This enables the model to capture localized pollution 
variations effectively, especially in urban environments with heterogeneous spatial patterns particularly 
along high-traffic arterials where exposure hotspots emerge near intersections, idling zones, and freight 
hubs.

To represent the dynamic nature of commuter trajectories, we define a time-evolving graph that adapts 
to real-world mobility patterns.
Formally, we define a graph: $ G = (V, E) $ where $( V )$ represents the set of spatial segments or clusters (nodes) generated via either fixed-size segmentation or density-based clustering (DBSCAN) of the mobile sensing data stream. The edges \( E \) encode spatial relationships, computed using geographical proximity or actual travel distances between nodes to better reflect human exposure patterns.
For forecasting, consider the following notations: At each time instance \( t \), the historical PM\textsubscript{2.5} concentration values for all \( n \) nodes (i.e., segments or clusters) are given by:
\begin{equation}
	x_t^{\text{PM}{2.5}} = \left( x_{(t-r)}, x_{(t-r+1)}, \ldots, x_t \right)
\end{equation}

where \( x_{(t-l)} \in \mathbb{R}^n \) denotes the PM\textsubscript{2.5} concentrations at all nodes at the \( (t-l)^\text{th} \) time step, and \( r \) is the lookback window length or historical length.

Additionally, we incorporate auxiliary feature sequences such as meteorological variables, represented by:
\begin{equation}
	x^{\text{feature}}_t = \left( \tilde{x}_{t - r}, \dots, \tilde{x}_{t- 1},\tilde{x}_t, \dots, \tilde{x}_{t + u} \right)
\end{equation}

where \( \tilde{x}_{t - l} \in \mathbb{R}^{n \times d} \) denotes the feature vector for all \( n \) nodes at time step \( t - l \), and \( u \) indicates the forecast horizon or the number of future time steps to predict.

\noindent\textbf{Objective:}

The objective is to learn a function \( f_W \), parameterized by \( W \), such that:
\[
\left( x^{\text{PM}{2.5}}_{t}, x^{\text{PM}{2.5}}_{t+1}, \dots, x^{\text{PM}{2.5}}_{t+u} \right) = f_W\left( x^{\text{PM}_{2.5}}_t, x^{\text{feature}}_t \right)
\]

Here, \( f_W \) corresponds to the proposed Spatially Attentive Cluster-based Graph Neural Network (SA-GNN), producing the forecasted PM\textsubscript{2.5} concentrations for all spatial segments or clusters in the graph over future timestamps.

\begin{figure}[H]
	\centering
	\includegraphics[width=0.80\linewidth, height=4.5cm]{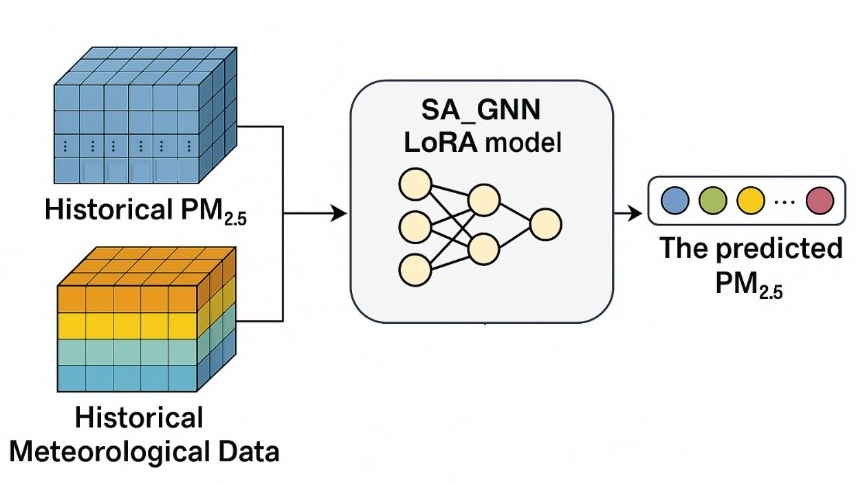}
	\caption{Overview of the SA-GNN-LoRA model architecture. The model ingests historical PM\textsubscript{2.5} and meteorological data to learn spatial and temporal patterns across clustered graph nodes, predicting future PM\textsubscript{2.5} concentrations.}
	\label{fig:sa-gnn-architecture}
\end{figure}

By modeling each spatial segment or cluster as a graph node, this formulation allows the model to capture both localized temporal patterns within individual clusters and spatial dependencies between clusters, providing high-resolution and personalized air quality forecasts for mobile sensing applications (see Figure~\ref{fig:sa-gnn-architecture}).
\vspace{-1em}
\section{Dataset Description and Preprocessing}\label{sec:experimental}

To enable fine-grained, mobility-aware PM2.5 forecasting, a high-resolution mobile sensing dataset was 
collected along a 14 km urban arterial corridor in Surat, India, capturing real-time commuter exposure 
under heterogeneous traffic and land-use conditions.
This section outlines the dataset, data collection methodology, and preprocessing steps under
taken before model development. Section~\ref{sec:4.1} introduces the study area and describes the spatial
characteristics of the dataset. Section~\ref{sec:4.1.1} details the mobile sensing platform and data acquisition
procedures, including temporal coverage, sensor calibration, and synchronization protocols. Section~\ref{sec:4.1.2} discusses data preprocessing operations such as normalization, missing value imputation, and
node definition for spatial graph construction. It also presents descriptive and statistical analyses of
PM2.5 concentration patterns, highlighting temporal, seasonal, and spatial variations across different
land-use categories. Finally, the section concludes by summarizing key findings and their implications
for the subsequent modeling framework.

\subsection{Study Corridor}\label{sec:4.1}

\begin{figure}[H]
	\centering
	\begin{subfigure}[b]{0.45\textwidth}
		\centering
		\includegraphics[width=\linewidth, height=3.5cm]{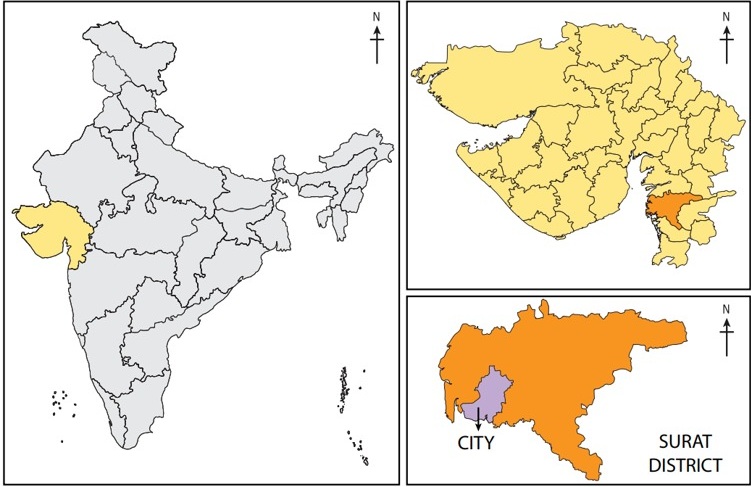}
		\caption*{\textbf{(a)}}
		\label{fig:image1}
	\end{subfigure}
	\hfill
	\begin{subfigure}[b]{0.51\textwidth}
		\centering
		\includegraphics[width=\linewidth, height=3.5cm]{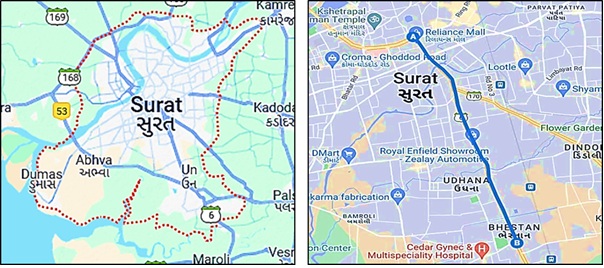}
		\caption*{\textbf{(b)}}
		\label{fig:image2}
	\end{subfigure}
	\caption{(a) Map of Surat, Gujarat, India and (b) The boundary of Surat and study route.}
	\label{fig:surat}
\end{figure}

The empirical foundation of this research lies in Surat, Gujarat, one of India's most industrialized and rapidly urbanizing cities. Surat is
globally recognized as a hub for textile manufacturing, diamond
polishing, and heavy industries, while simultaneously grappling with
rapid population growth, expansion of built-up areas, and growing
vehicular density. These dynamics exert tremendous pressure on the
city's transport infrastructure and contribute to its deteriorating air
quality. Like many Indian cities, Surat faces the dual challenge of
traffic congestion and air quality degradation, which interact in
complex ways to shape commuter exposure to fine particulate matter
( PM\textsubscript{2.5}) whose geographic layout is depicted in Fig.~\ref{fig:surat}.

The selected study corridor is a 14 km bidirectional arterial link
between Udhna and Bhestan (Fig.~\ref{fig:surat}A), representing one of the busiest commuter and freight routes in the city. This corridor provides a cross-sectional
representation of Surat's urban form, connecting the older Udhna area
characterized by textile clusters and wholesale markets with the
expanding Bhestan periphery, where residential colonies and transport
facilities are rapidly emerging. The corridor is flanked by a
heterogeneous mix of land uses, including dense residential settlements,
wholesale and retail commercial establishments, major industrial
estates, freight yards, and bus terminals. Such diversity creates
localized emission hotspots, particularly near signalized intersections,
industrial frontages, and curbside commercial strips, where stop and go
driving and idling amplify pollutant accumulation.

Traffic volumes along the corridor are among the highest in Surat, with
heterogeneous traffic streams comprising two-wheelers, three-wheelers,
cars, buses, and heavy trucks competing for limited right-of-way.
Morning and evening rush hours are characterized by severe congestion,
with average speeds frequently falling below 15 km/h near major
intersections. This results in idling emissions, repeated acceleration
and braking, and resuspension of road dust, all of which aggravate  PM\textsubscript{2.5}
levels. Meteorological conditions further exacerbate the problem:
Surat's coastal climate is marked by high relative humidity, and during
winter, low atmospheric mixing heights trap pollutants near the surface,
leading to episodic high concentrations. These combined conditions make
the corridor an archetype of the broader transport-environment
challenges faced by rapidly urbanizing Indian cities.

Along its 14 km alignment, the corridor includes seven major
intersections, which serve as both traffic bottlenecks and land-use
transition points. For clarity, they are denoted as Intersection 1
through Intersection 7. Intersection 1 functions as the gateway to the
corridor, with high commuter inflows and mixed residential-commercial
activity. Intersection 2 lies within a residential-commercial transition
zone prone to localized congestion. Intersection 3 is a multi-arm
arterial junction and one of the most congested locations along the
route, shaped by heavy modal mixing and pedestrian activity.
Intersection 4 is adjacent to a large industrial cluster, strongly
influenced by freight traffic. Intersection 5 is a signalized roundabout
combining industrial and commercial activity, with long queues of idling
vehicles. Intersection 6 lies at a residential-industrial interface,
where congestion is moderate but persistent. Finally, Intersection 7
serves as the terminal junction, adjoining regional transport facilities
and newly built residential colonies.

These intersections form critical anchor points for understanding
exposure heterogeneity. Their distribution allows for systematic
examination of how land-use shifts and traffic intensity combine to
influence  PM\textsubscript{2.5} variability. These
locations consistently coincide with traffic bottlenecks, reinforcing
the link between idling emissions, queuing traffic, and pollutant
accumulation.
The choice of this study area is therefore motivated by three
interrelated factors. First, it represents an archetypal congested
Indian arterial that integrates residential, industrial, and commercial
land uses, making it ideal for examining how heterogeneity drives
pollutant concentrations. Second, it experiences recurrent congestion
and modal interaction, offering a natural laboratory for exploring the
dynamics of transport-pollution interactions at fine temporal and
spatial scales. Third, the corridor is of strategic policy relevance: as
a key commuter link and freight connector, evidence from this study can
inform mitigation strategies not only for Surat but also for other
rapidly urbanising cities are facing similar challenges.
\subsubsection{Mobile Sensing and Data Collection}\label{sec:4.1.1}
To replicate real-world commuter exposure, a motorized two-wheeler (MTW)-mounted sensing platform was 
deployed, collecting high-frequency data across 53 bidirectional runs. To capture commuter-level exposure, a 
motorized two wheeler (MTW)-mounted mobile sensing platform was deployed to replicate real-world travel 
conditions.

\begin{figure}[H]
	\centering
	\includegraphics[width=0.5\linewidth]{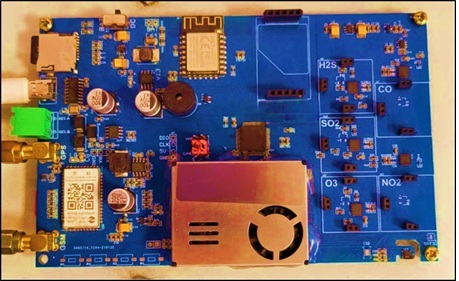}
	\caption{MTW mounted mobile monitoring PM\textsubscript{2.5} sensors, meteorological probes, and GPS, designed to simulate commuter exposure}
	\label{fig:sensor}
\end{figure}
\paragraph{\textbf{Mobile Sensing Platform:}}
To capture fine-scale variations in air quality, a bike-mounted sensing 
platform was developed to replicate commuter-level exposure along the study corridor. The platform was 
equipped with a calibrated optical particle counter for PM2.5, sensor as shown in Fig.~\ref{fig:sensor}., supported by 
meteorological probes measuring temperature, relative humidity and a 
GPS unit provided continuous spatial referencing, ensuring accurate mapping of pollutant 
concentrations. Data were logged at a 3-second interval, chosen to balance temporal resolution with sensor 
stability. The campaign spanned diurnal, weekly, and seasonal cycles to capture full mobility and 
meteorological variability, yielding over 50,000 synchronized observations. 

\paragraph{\textbf{Monitoring Protocol:}}
The monitoring campaign consisted of 53 repeated runs across the 14 km Udhna--Bhestan corridor. Runs were scheduled during the morning
(06:00--10:00), afternoon (12:00--16:00), and evening (18:00--21:00)
periods to capture diurnal variability, and were spread across
post-monsoon, winter, and pre-monsoon seasons to reflect seasonal
heterogeneity. Each run covered both forward and reverse directions,
ensuring bidirectional representation of traffic dynamics. The campaign produced a rich dataset containing  PM\textsubscript{2.5} concentrations, meteorological attributes, land-use fractions, and spatial coordinates, establishing one of the most detailed mobile monitoring datasets for an Indian urban corridor.
Data quality was ensured through rigorous imputation and synchronization, enabling robust graph 
node construction. 

\paragraph{\textbf{Data Completeness and Missingness:}}\label{sec:4.1.2}
Data gaps (\(<5\%\)) from sensor warm-up, GPS loss, and logging failures were addressed using a two-step imputation: short gaps (\(\le 15\) s) were filled via cross-run averaging, while longer gaps or run-end losses used adjacent-day averaging. This preserved diurnal patterns, temporal continuity, and extreme values for forecasting. 

\paragraph{\textbf{Synchronization and Node Definition:}}
All data streams were synchronized to a 3-second temporal grid. GPS and pollutant data were interpolated, while meteorological and land-use variables were aggregated by segment. The route was divided into 200 m segments, each serving as a graph node, and DBSCAN clustering identified high-exposure hotspots. These node definitions enabled construction of the Spatially Attentive Graph Neural Network (SA-GNN) for subsequent modeling.

\subsubsection{Data Preprocessing}\label{sec:4.1.2}
\begin{figure}[htbp!]
	\centering
	\begin{subfigure}[b]{0.45\textwidth}
		\centering
		\includegraphics[width=\textwidth, height=3cm]{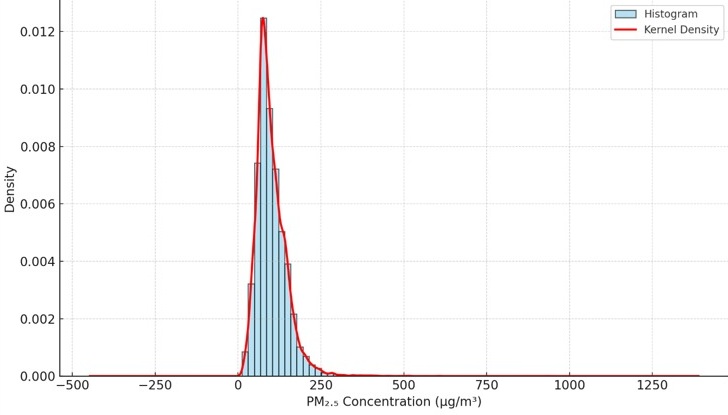}
		\caption{}
	\end{subfigure}
	\hfill
	\begin{subfigure}[b]{0.45\textwidth}
		\centering
		\includegraphics[width=\textwidth, height=3cm]{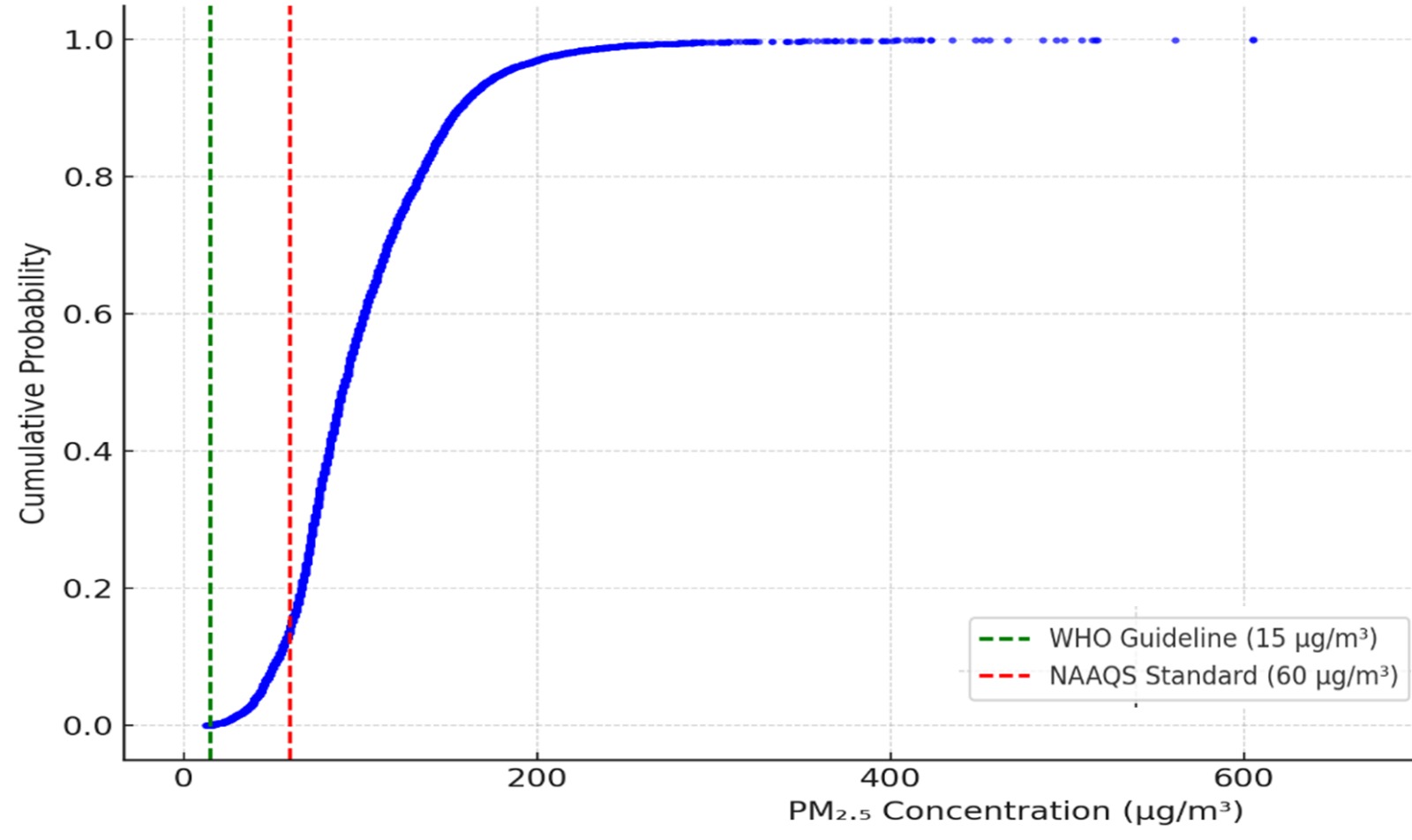}
		\caption{}
	\end{subfigure}
	\caption{a) Histogram of PM\textsubscript{2.5} concentrations with kernel density overlay. The distribution is skewed with long tails, confirming episodic extremes.
		and b) Empirical cumulative distribution function (ECDF). Nearly 70\% of observations exceed WHO standards, and ~40\% exceed NAAQS.}
	\label{fig:fig4}
\end{figure}

To prepare the mobility dataset for spatiotemporal graph learning, normalization and feature 
engineering were applied to ensure model stability and interpretability. 
\vspace{-1.5em}
\paragraph{\textbf{Data Normalization}}
To prepare the dataset for modeling, all variables were normalized using
z-score transformation, ensuring comparability across features. The
target variable  PM\textsubscript{2.5} was also normalized during training, which explains why its LIME importance values appear relatively lower compared to exogenous predictors- a preprocessing effect clarified in the results section. Normalization ensured that variables with different scales did not bias the learning process while maintaining interpretability of pollutant trends in absolute terms.
\vspace{-1em}
\paragraph{\textbf{Descriptive Statistics of PM\textsubscript{2.5} Concentration Patterns}}
Descriptive analysis reveals chronic, traffic
driven PM2.5 exposure with strong diurnal, weekly, and spatial heterogeneity validating the need for 
adaptive, corridor-scale forecasting along the 14 km corridor. The mean concentration was 100.5
µg/m³, with a median of 91 µg/m³ and a standard deviation of 46.0 µg/m³.
Concentrations ranged from 12 µg/m³ to extreme values as high as 930
µg/m³, indicating the occurrence of severe episodic spikes. The
interquartile range (Q1-Q3) was 70-124 µg/m³, well above both the WHO
24-hour guideline (15 µg/m³) and the Indian NAAQS threshold (60 µg/m³).
These statistics reveal a right-skewed distribution with a heavy upper
tail, typical of urban corridors dominated by dense traffic and
industrial activity. Notably, over 70\% of the observations exceeded WHO
guidelines, and nearly 40\% exceeded national standards, highlighting
chronic commuter exposure as illustrated in Fig. ~\ref{fig:fig4}.

\begin{figure}[H]
	\centering
	\includegraphics[width=0.6\linewidth, height=3cm]{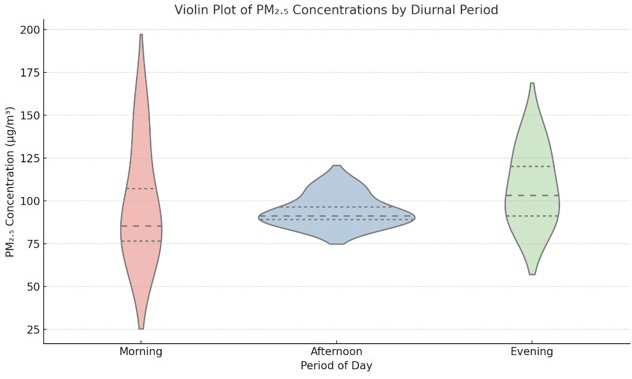}
	\caption{Variation of PM2.5 concentration for period of day (Morning, Afternoon,Evening).}
	\label{fig:variation_pm25}
\end{figure}

\paragraph{\textbf{Temporal Variations}}
Temporal disaggregation confirms systematic daily patterns. Morning runs
(06:00--10:00) averaged 101.7 µg/m³, and evening runs (18:00--21:00)
were even higher at 106.8 µg/m³, both corresponding to peak-hour traffic
and suppressed boundary layer conditions. In contrast, afternoon periods
(12:00--16:00) showed lower means of 93.3 µg/m³, consistent with
enhanced dispersion from convective mixing. Boxplots by time of day
(Fig. ~\ref{fig:variation_pm25}) reinforce the bimodal nature of exposure, characteristic of
commuter corridors.

\vspace{-2em}
\paragraph{\textbf{Weekday-Weekend and Day-of-Week Patterns:}}
Weekly stratification highlights pronounced contrasts (Table ~\ref{tab:weekday}).
Weekdays averaged 108.4 µg/m³, nearly 40\% higher than weekends (64.2
µg/m³), consistent with reduced traffic and partial industrial shutdowns
on weekends. Among weekdays, Tuesday and Thursday showed the highest
averages (111.9 and 119.7 µg/m³), with maxima exceeding 600 µg/m³,
reflecting mid-week freight and industrial activity. By contrast,
Saturday and Sunday averages were much lower (63.1 and 65.9 µg/m³).

\begin{figure}[H]
	\centering
	\includegraphics[width=0.85\linewidth, height=4cm]{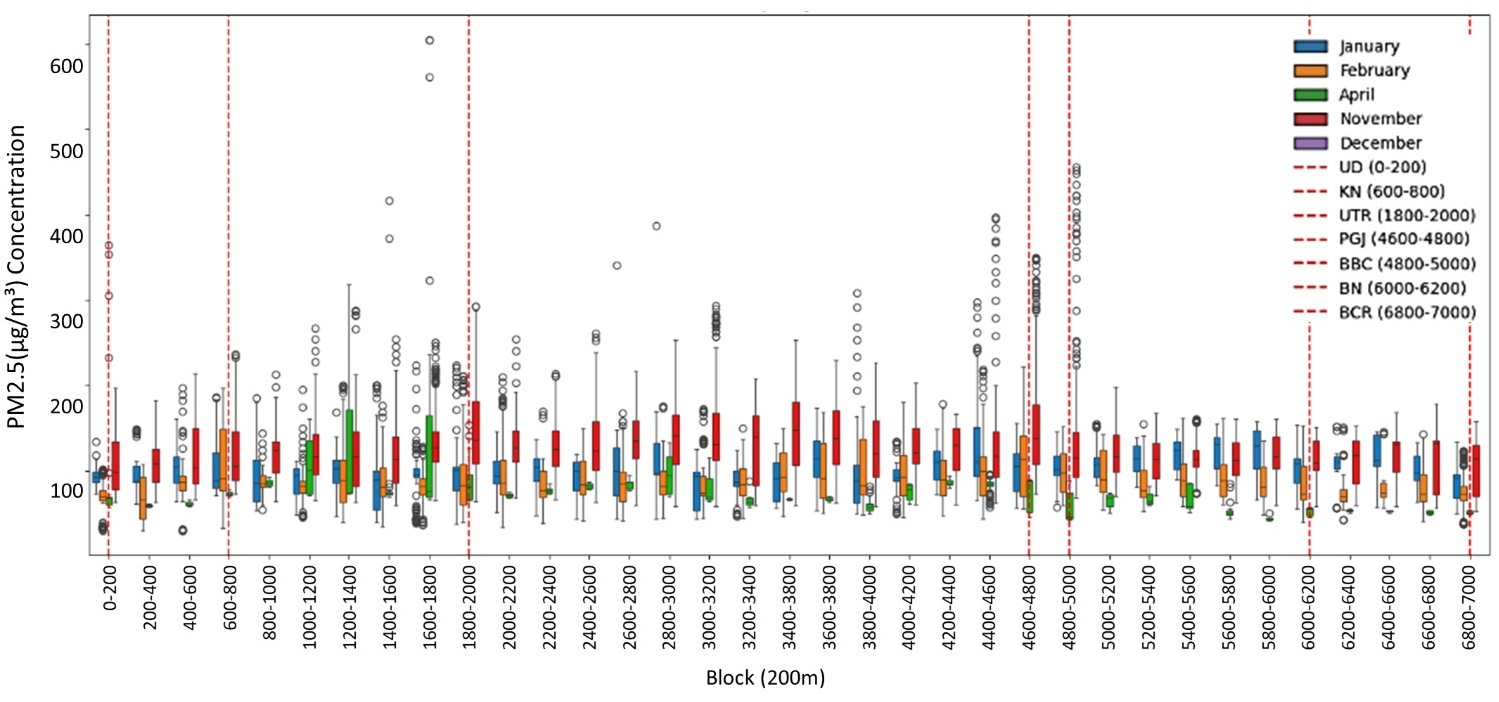}
	\caption{Monthly block-wise boxplots of PM\textsubscript{2.5} concentrations. Winter months are most polluted, April shows the cleanest conditions.}
	\label{fig:seasonal_boxplot}
\end{figure}

\begin{table}[h!]
	\centering
	\caption{PM$_{2.5}$ descriptive statistics by day of week ($\mu g/m^3$)}
	\begin{tabular}{lccc}
		
		\hline
		\textbf{Day} & \textbf{Mean (PM$_{2.5}$)} & \textbf{Std. Dev. (PM$_{2.5}$)} & \textbf{Max (PM$_{2.5}$)} \\
		\hline
		Monday    & 94.4  & 20.8 & 162 \\
		\hline
		Tuesday   & 111.9 & 41.4 & 409 \\
		\hline
		Wednesday & 97.1  & 38.6 & 517 \\
		\hline
		Thursday  & 119.7 & 49.4 & 605 \\
		\hline
		Friday    & 55.7  & 15.1 & 106 \\
		\hline
		Saturday  & 63.1  & 26.0 & 253 \\
		\hline
		Sunday    & 65.9  & 36.4 & 241 \\
		\hline
		\label{tab:weekday}
	\end{tabular}
\end{table}
\paragraph{\textbf{Monthly and Seasonal Trends:}}
Seasonal disaggregation highlights winter dominance(Fig.~\ref{fig:seasonal_boxplot}). January--February
recorded the highest means (\textgreater120 µg/m³), driven by
temperature inversions and low mixing heights. Post-monsoon months
(October--November) also showed elevated levels due to stagnant
meteorology and pollutant accumulation. In contrast, April and May
exhibited the lowest averages (\textasciitilde85 µg/m³), benefitting
from stronger winds and higher convective activity. These shifts
emphasize the strong interplay between emissions and meteorology.

\begin{figure}[H]
	\centering
	\begin{subfigure}[b]{0.48\textwidth}
		\centering
		\includegraphics[width=\linewidth, height=4cm]{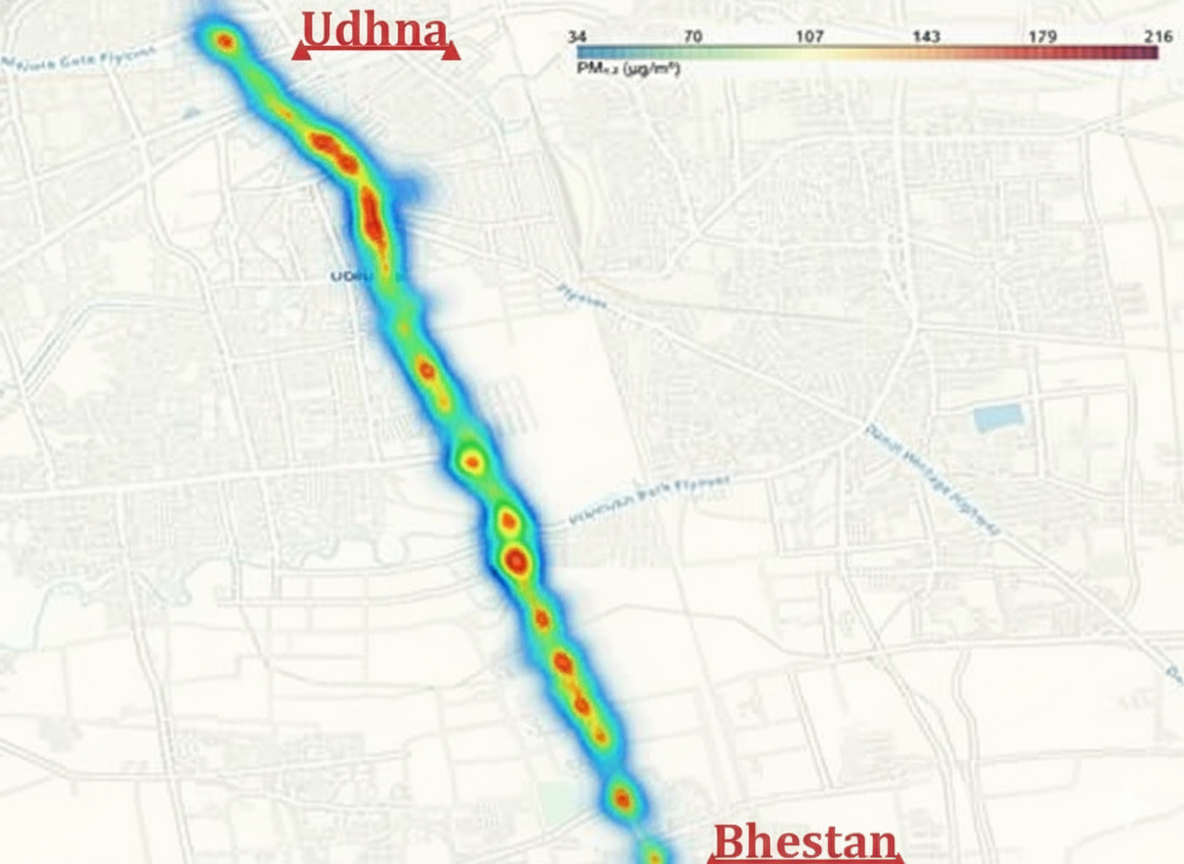}
		\label{fig:image1}
	\end{subfigure}
	\hfill
	\begin{subfigure}[b]{0.48\textwidth}
		\centering
		\includegraphics[width=\linewidth, height=4cm]{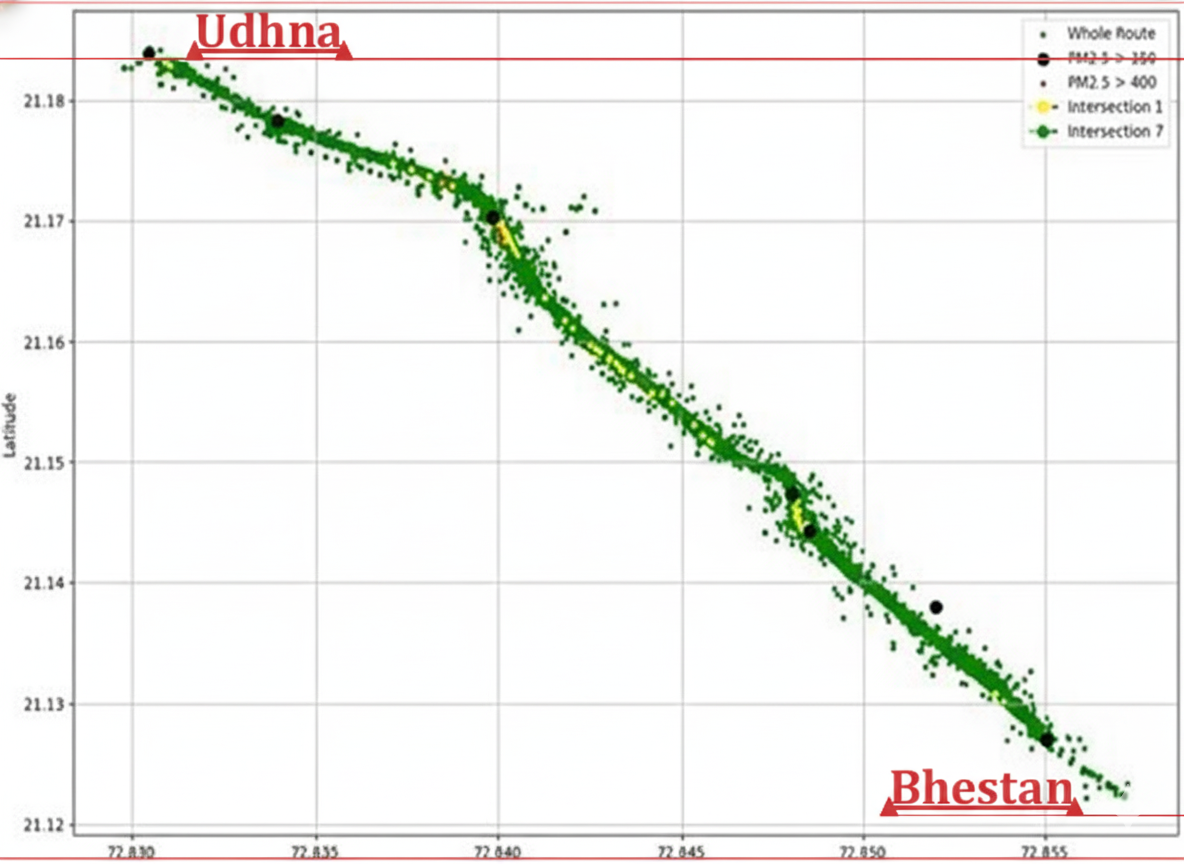}
		\label{fig:image2}
	\end{subfigure}
	\caption{Spatial heatmaps along the 14 km corridor. Intersections 3–5 emerge as consistent hotspots across time periods.}
	\label{fig:fig7}
\end{figure}

\begin{figure}[H]
	\centering
	\includegraphics[width=0.7\linewidth, height=5cm]{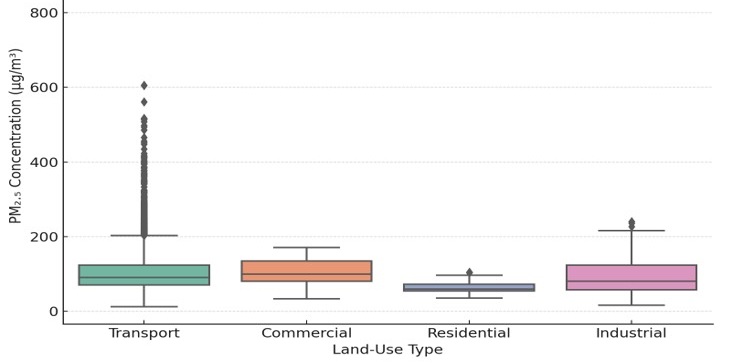}
	\caption{Boxplots of PM\textsubscript{2.5} by land-use type. Industrial $>$ Commercial $>$ Transport $>$ Residential, with larger variability in industrial zones.}
	\label{fig:boxp}
\end{figure}

\vspace{-2em}
\paragraph{\textbf{Spatial Heterogeneity and Land-Use Influence:}}
Spatial aggregation at 200 m resolution confirmed intersection-driven
hotspots (Table. ~\ref{tab:intersection}). The 4600-4800 m segment (Intersection 4-5 stretch) recorded
the highest mean concentration of 119.8 µg/m³, followed by the 1800-2000m segment (Intersection 3) with 117.2 µg/m³, and its approach 1600-1800m with 105.8 µg/m³. The 4800-5000 m segment (Intersection 5) also showed sustained high levels (105.6 µg/m³). These hotspots coincide with heavy congestion, idling queues, and industrial frontage as illustrated in the spatial heatmaps (Fig.~\ref{fig:fig7})..
By contrast, residential stretches averaged \textasciitilde61 µg/m³,
which, while lower, still far exceed health-based standards.Land-use stratification reinforced these findings (Table ~\ref{tab:landuse}).
Industrial areas recorded an average PM\textsubscript{2.5} concentration of 95.1 µg/m³, followed by commercial zones at 82.7 µg/m³ and transport corridors at 78.5 µg/m³, while residential areas exhibited comparatively lower levels. The corresponding distribution across land-use categories averaged 61.3 µg/m³ is illustrated in Fig.~\ref{fig:boxp}.
\begin{table}[h!]
	\centering
	\begin{minipage}{0.48\textwidth}
		\centering
		\caption{Major intersections}
		\begin{tabular}{lccl}
			\hline
			\textbf{Segment (m)} & \textbf{Mean} & \textbf{Max} & \textbf{Intersection} \\
			\hline
			1600--1800 & 105.8 & 605 & Intersection 3 approach \\
			1800--2000 & 117.2 & 508 & Intersection 3 \\
			4600--4800 & 119.8 & 282 & Intersection 4--5 stretch \\
			4800--5000 & 105.6 & -- & Intersection 5 \\
			\hline
		\end{tabular}
		
		\label{tab:intersection}
	\end{minipage}
	\hfill
	\begin{minipage}{0.48\textwidth}
		\centering
		\caption{By land-use}
		\hspace{2cm}
		\begin{tabular}{lcc}
			\hline
			\textbf{Land-Use} & \textbf{Mean} & \textbf{Max} \\
			\hline
			Industrial & 95.1 & 305 \\
			Commercial & 82.7 & 276 \\
			Transport  & 78.5 & 251 \\
			Residential & 61.3 & 188 \\
			\hline
		\end{tabular}
		
		\label{tab:landuse}
	\end{minipage}
\end{table}

\paragraph{\textbf{Summary and Implications}}
The descriptive statistics establish several critical insights. First,
overall exposure is alarmingly high, with mean concentrations nearly
seven times WHO guidelines and maxima approaching 605 µg/m³. Second,
temporal heterogeneity is pronounced, marked by diurnal peaks,
weekday-weekend contrasts, and seasonal cycles. Third, spatial
variability is substantial, with industrial and transport-dominated
segments acting as persistent hotspots.

These findings directly address reviewer concerns by clarifying dataset
structure, completeness, and variability. They also justify the adoption
of advanced spatiotemporal frameworks. Traditional station-based or
simple time-series approaches cannot adequately capture the skewed
distribution, intersection-level hotspots, and land-use dependencies
observed here. The observed patterns motivate the use of Spatially
Attentive Graph Neural Networks (SA-GNNs), which combine
cluster-specific GRUs for localized temporal learning with graph
attention mechanisms for capturing cross-segment dependencies. Thus, the
descriptive statistics not only provide empirical grounding but also
reinforce the methodological significance of the proposed modeling
framework.

\vspace{-.5em}
\subsection{Hyperparameter Setting}\label{sub:hypermeter}



The SA-GNN-LoRA model is trained to minimize the Mean Squared Error (MSE) loss between predicted and actual PM$_{2.5}$ concentrations using backpropagation. Model parameters are initialized with a random normal distribution and optimized using the AdamW optimizer, which combines adaptive learning rates with decoupled weight decay for improved generalization. A Cosine Annealing Learning Rate Scheduler is employed to adjust the learning rate dynamically across epochs, aiding convergence and helping to escape local minima.GRU cells within both the temporal module and the cluster-wise spatiotemporal block are configured with a hidden dimension of 128 to capture complex temporal dependencies. For regularization, an L1 penalty is applied to control model complexity and mitigate overfitting. L2 regularization is avoided due to its tendency to cause divergence between training and validation losses, resulting in premature early stopping.The model is trained for 100 epochs using a 5-fold cross-validation strategy. The learning rate is set to 0.0001, and the weight decay is 0.00001. Hyperparameters are selected through empirical trial-and-error experimentation, and all reported results are based on this data split.
\subsection{Clustering Monitoring Station}
\noindent To effectively capture localized spatiotemporal dynamics of PM$_{2.5}$ variations, a hybrid strategy combining clustering and segmentation was employed to define virtual monitoring stations along the study route. First, the Density-Based Spatial Clustering of Applications with Noise (DBSCAN) algorithm was applied to a feature space consisting of geographic coordinates and average PM$_{2.5}$ concentrations. This density-based method successfully identified 18 coherent clusters of spatially adjacent data points with similar pollution levels, while also filtering out noise.

\begin{figure}[H]
	\centering
	\begin{minipage}[t][4cm][t]{0.47\textwidth}  
		\centering
		\includegraphics[width=\textwidth,height=5.0cm]{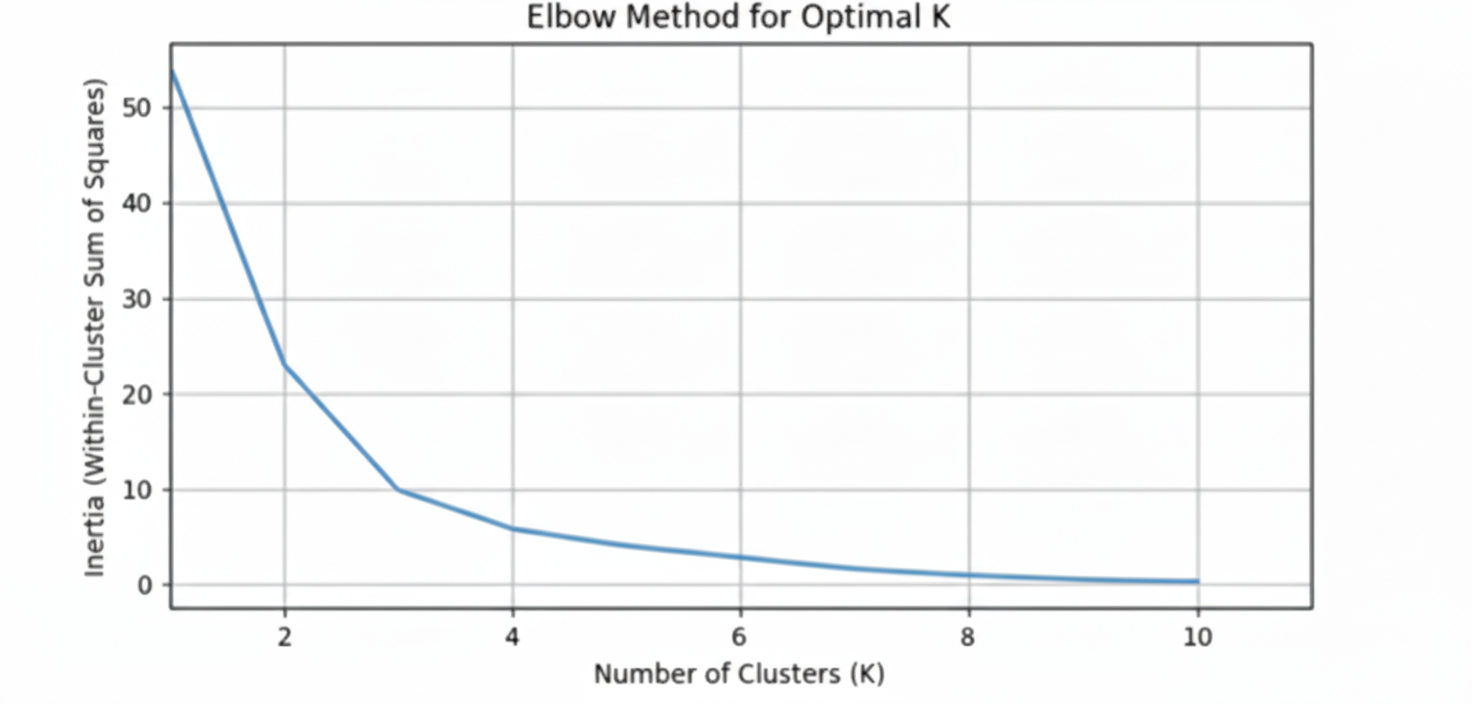}
		\subcaption*{(a) Elbow Method plot used to determine the optimal number of KMeans clusters ($K=3$), also confirmed using the \texttt{KneeLocator} algorithm.}
		\label{fig:elbow}
	\end{minipage}
	\begin{minipage}[t][7cm][t]{0.47\textwidth}  
		\centering
		\includegraphics[width=\textwidth,height=5.0cm]{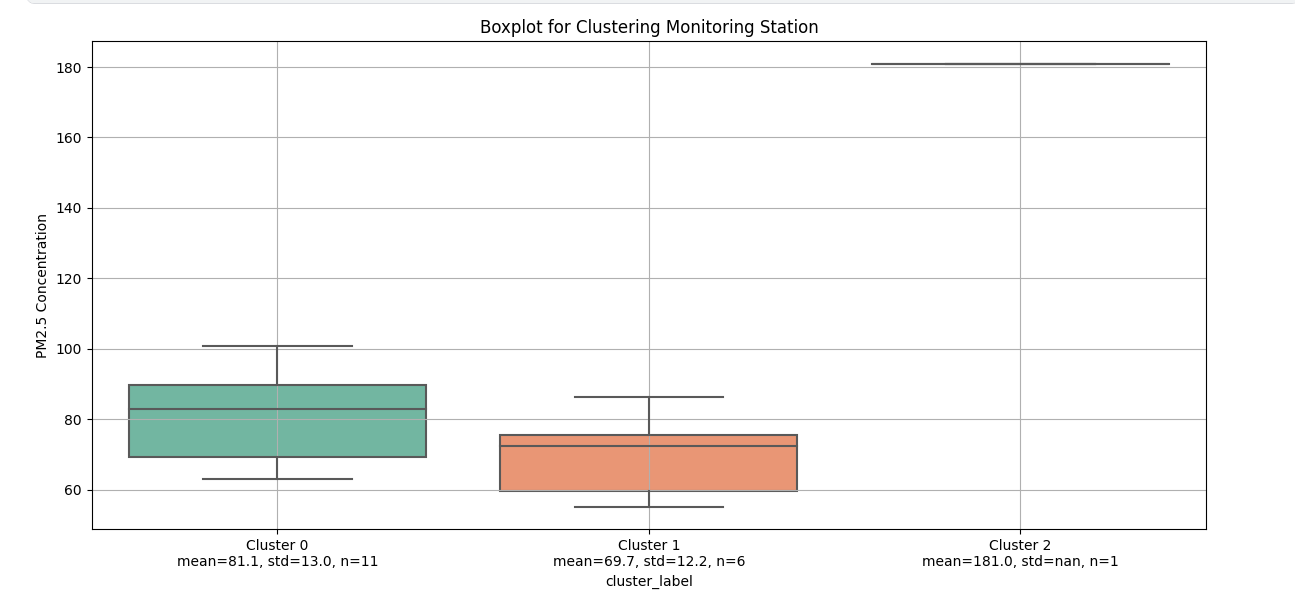}
		\subcaption*{subfigure}{(b)Cluster-wise Boxplot for PM\textsubscript{2.5} Concentration.}
		\label{fig:boxplot}
	\end{minipage}
	
	\caption{Clustering results visualization: (a) Elbow method and (b) PM\textsubscript{2.5} cluster-wise boxplot.}
	\label{fig:combined_elbow_boxplot}
\end{figure}

\vspace{-2em}
To ensure complete spatial coverage particularly in areas with sparse sampling or uneven sensor distribution DBSCAN was complemented with uniform grid segmentation, dividing the urban route into fixed 200-meter intervals. PM$_{2.5}$ readings within each segment were aggregated to form representative values, ensuring consistent spatial granularity.

To uncover higher-level pollution zones, K-Means clustering was applied to the centroids of the DBSCAN clusters using normalized latitude, longitude, and average PM$_{2.5}$. The optimal number of clusters ($K=3$) was determined using the Elbow Method, with the \texttt{Knee Locator} algorithm selecting the ideal cutoff, as shown in Figure~\ref{fig:combined_elbow_boxplot}.

The resulting clusters revealed distinct pollution profiles: Cluster 0 (11 stations) had a mean PM$_{2.5}$ concentration of 81.1\,µg/m$^3$, Cluster 1 (6 stations) averaged 69.7\,µg/m$^3$, and Cluster 2 (1 station) exhibited a significantly higher mean of 181.0\,µg/m$^3$, as shown in Figure~\ref{fig:combined_elbow_boxplot}, marking it as the most polluted region. These cluster definitions are integral to the SA-GNN framework, guiding the cluster-specific GRU modules to learn localized temporal patterns associated with different pollution regimes.

\vspace{-1em}
\subsection{Feature Correlation and Selection}\label{sec:featureselection}
\label{sec:featureselection}
\vspace{0.5em}

\noindent To identify the most relevant features for PM$_{2.5}$ prediction, a correlation-based selection strategy was employed, combining three statistical techniques: Pearson correlation, Spearman correlation, and Gray Relational Analysis (GRA). These methods quantify linear, monotonic, and geometric similarity relationships between PM$_{2.5}$ and other candidate features.

\begin{figure}[!htbp]
	\centering
	\includegraphics[width=\textwidth]{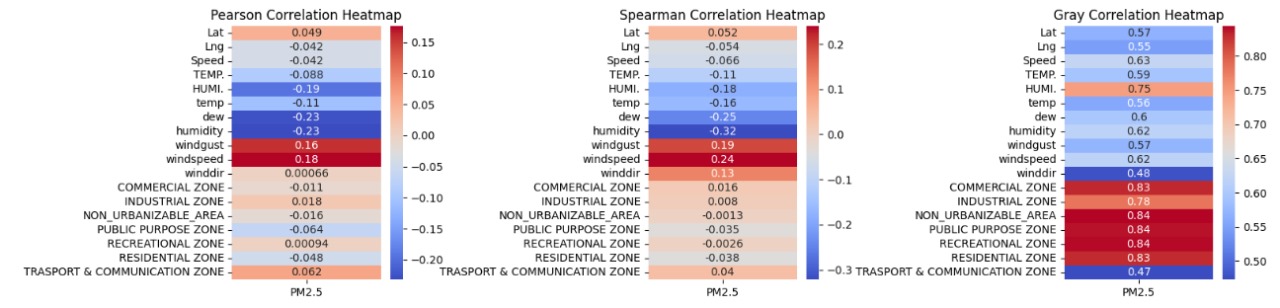}
	\caption{ Correlation heat maps based on Pearson, Spearman, and Grey correlation methods for the selected features.}
	\label{fig:correlation_heatmaps}
\end{figure}

\textbf{Pearson Correlation:}
Pearson correlation measures the linear relationship between two continuous variables. The coefficient $r_{xy}$ is calculated as:
\vspace{-1em}
\begin{equation}
	r_{xy} = \frac{\sum_{i=1}^n (x_i - \bar{x})(y_i - \bar{y})}{\sqrt{\sum_{i=1}^n (x_i - \bar{x})^2} \sqrt{\sum_{i=1}^n (y_i - \bar{y})^2}}
\end{equation}
where $x$ and $y$ are the two variables, and $\bar{x}$ and $\bar{y}$ are their respective means.
\textbf{Spearman Correlation:}
Spearman correlation is a non-parametric measure of rank correlation. It assesses how well the relationship between two variables can be described using a monotonic function. It is given by:
\begin{equation}
	q_s = \frac{\sum_{i=1}^n (R_i - \bar{R})(S_i - \bar{S})}{\sqrt{\sum_{i=1}^n (R_i - \bar{R})^2 \sum_{i=1}^n (S_i - \bar{S})^2}}
\end{equation}
where $R_i$ and $S_i$ are the ranks of the $i$-th observations of $X$ and $Y$.
\textbf{Gray Relational Analysis (GRA) :}
Gray Relational Analysis (GRA) evaluates geometric similarity between two sequences. Given a reference sequence $X_0$ (e.g., PM$_{2.5}$) and a comparison sequence $X_i$ (a candidate feature), the gray relational coefficient $\gamma(x_i, y_i)$ is computed as:
\begin{equation}
	\begin{aligned}
		\gamma(x_i, y_i) &= \frac{\min |x_i - y_i| + \rho \, \max |x_i - y_i|}{|x_i - y_i| + \rho \, \max |x_i - y_i|}, \\
		\xi &= \frac{1}{n} \sum_{i=1}^{n} \gamma(x_i, y_i)
	\end{aligned}
\end{equation}
where $\rho$ is the distinguishing coefficient, typically set to 0.5.
This multi-metric approach enabled robust selection of relevant variables, as summarized in Figure  ~\ref{fig:correlation_heatmaps}. Meteorological features such as temperature, humidity, wind speed, wind gust, dew point, and wind direction were selected. Although wind direction showed weaker Pearson and Spearman correlations, it was retained due to its directional influence on pollutant transport.
In addition, contextual variables such as vehicle speed and 
zoning attributes including COMMERCIAL ZONE, INDUSTRIAL ZONE, RESIDENTIAL ZONE, and 
PUBLIC PURPOSE ZONE were incorporated to account for spatial heterogeneity in emissions across urban 
environments.
\section{Network Architecture}\label{sec:architecture}
\begin{figure}[H]
	\centering
	\includegraphics[width=1.0\textwidth, height=0.25\textheight]{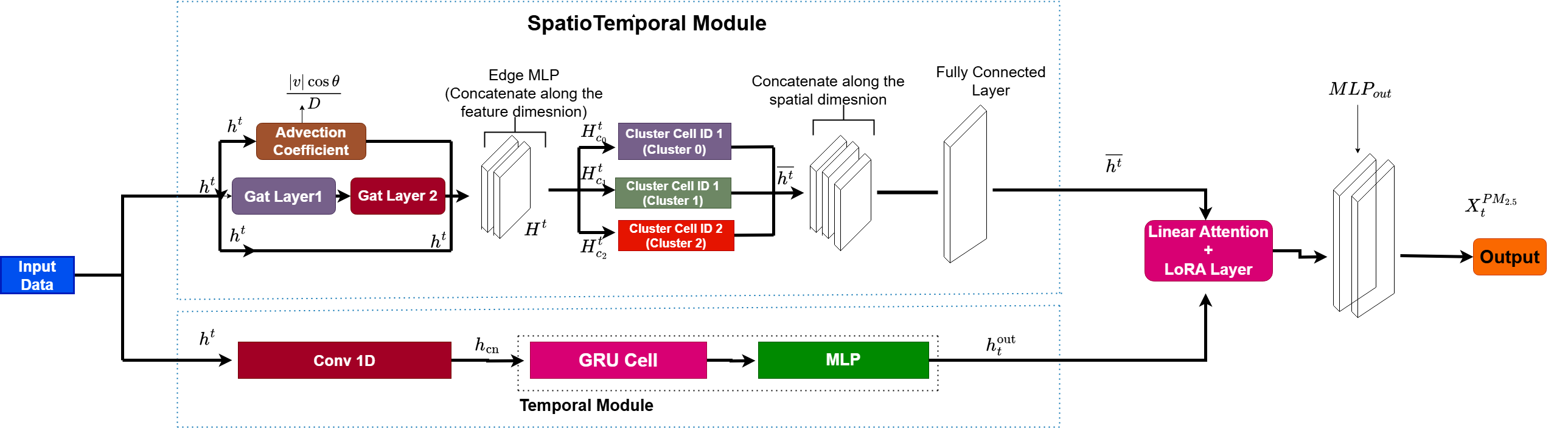}
	\caption{Overview of the SA-GNN-LoRA architecture, showing the spatiotemporal and temporal modules, pollutant transport modeling, clustering, and final attention-based prediction.}
	\label{fig:saganlora}
\end{figure}
\noindent The proposed SA-GNN architecture effectively captures the complex dynamics inherent in PM$_{2.5}$ data by jointly modeling spatial and temporal dependencies. The Spatiotemporal Module leverages graph-based attention mechanisms to dynamically learn spatial correlations among monitoring stations, enabling the model to account for the geographical distribution and interactions of pollution sources. Simultaneously, the Temporal Module incorporates recurrent and convolutional structures to capture temporal patterns, including daily and seasonal variations, as well as abrupt changes in pollution levels over time. 

The Low-Rank Adaptation component enhances model efficiency by reducing parameter redundancy while preserving representational capacity, thereby improving generalization and accelerating convergence during training. This modular design enables SA-GNN to comprehensively and efficiently model the spatiotemporal complexity of air quality data, resulting in superior predictive performance. Figure ~\ref{fig:saganlora} illustrates the end-to-end architecture of the model, depicting the flow of information through both modules and highlighting how spatiotemporal features are fused and refined.

\subsection{Spatiotemporal Module}

\noindent The Spatiotemporal module of the SA-GNN model serves as a unified recurrent unit designed for multi-step forecasting of PM$_{2.5}$ concentrations. The underlying spatial structure is modelled as a directed graph $G=(V, E)$, where $V$ denotes the set of stations, and $E$ represents the edge connections characterized by pollutant transport properties. Each node is embedded with a feature vector combining real-time PM$_{2.5}$ measurements and various meteorological variables, including wind speed and direction, temperature, relative humidity, precipitation, and surface pressure. To capture pollutant movement between nodes, the model defines an \textit{advection coefficient} $P^t_{ji}$, which quantifies the influence of wind dynamics from source node $j$ to sink node $i$ at time $t$, given by equation~\ref{eq:2}:
\begin{equation}
	P^t_{ji} = \text{ReLU} \left( \frac{|v| \cos(\theta)}{D} \right)
	\label{eq:2}
\end{equation}
Here, $v$ denotes wind speed at node $j$, $\theta$ is the angle of wind direction relative to the line joining nodes $j$ and $i$, and $D$ is the geographical distance between them. The ReLU function ensures non-negative flow representation, aligning with physical transport constraints. This advection coefficient, along with the node features $h^t_j$ and $h^t_i$, is passed into an edge-MLP to compute directional pollutant transport flow in equation~\ref{eq:3}. The resulting edge features $e^t_{j \rightarrow i}$ are aggregated by a node-MLP layer to form the spatial pollutant transport embedding at node $i$ in equation~\ref{eq:4}.
\begin{align}
	e^t_{j \rightarrow i} &= \Lambda([h^t_j, h^t_i, P^t_{ji}]) \label{eq:3} \\
	\zeta^t_i &= \Psi\Big( \sum_{j \in \mathcal{N}(i)} e^t_{j \rightarrow i} + e^t_{i \rightarrow j} \Big) \label{eq:4}
\end{align}

Once the transport-aware node features are obtained, the model applies a 
\textit{two-layer Graph Attention Network (GAT)} to capture complex and 
hierarchical spatial dependencies among stations. In the first GAT layer, 
each node feature is linearly projected using a shared weight matrix 
\( W^{(1)} \in \mathbb{R}^{F \times F'} \), where \(F\) and \(F'\) denote 
the input and hidden feature dimensions, respectively. A pairwise attention 
coefficient \( e^{(1)}_{ij} \) is computed as:
\begin{align}
	e^{(1)}_{ij} &= \text{LeakyReLU}\Big( {({\vec{a}^{(1)}})^{\top}} 
	[W^{(1)} h_i \, \| \, W^{(1)} h_j] \Big) \label{eq:5} \\
	\alpha^{(1)}_{ij} &= \text{softmax}_j(e^{(1)}_{ij}) =
	\frac{\exp(e^{(1)}_{ij})}{\sum_{k \in \mathcal{N}_i} 
		\exp(e^{(1)}_{ik})} \\
	\tilde{h}^{(1)}_i &= \sigma \Big( \sum_{j \in \mathcal{N}_i} 
	\alpha^{(1)}_{ij} W^{(1)} h_j \Big)
\end{align}
The output of the first GAT layer $\tilde{h}^{(1)}_i$ is then passed to 
a second GAT layer to refine spatial dependencies and capture higher-order 
neighborhood interactions:
\begin{align}
	e^{(2)}_{ij} &= \text{LeakyReLU}\Big( {\vec{a}^{(2)}}{^\top} 
	[W^{(2)} \tilde{h}^{(1)}_i \, \| \, W^{(2)} \tilde{h}^{(1)}_j] \Big) \\
	\alpha^{(2)}_{ij} &= \text{softmax}_j(e^{(2)}_{ij}) \\
	\tilde{h}^{(2)}_i &= \sigma \Big( \sum_{j \in \mathcal{N}_i} 
	\alpha^{(2)}_{ij} W^{(2)} \tilde{h}^{(1)}_j \Big) \label{eq:7}
\end{align}

Here, $\tilde{h}^{(2)}_i$ represents the final spatially attended node feature 
after two stacked GAT layers, which is subsequently concatenated with the 
transport embedding $\zeta^t_i$ and the original feature $h^t_i$ to form 
the fused node-level representation in equation.

\subsubsection{Temporal Module}
\vspace{1em}
\noindent The \textbf{Temporal Module} models the sequential dynamics of PM$_{2.5}$ concentrations. It begins with a 1D Convolutional Neural Network (CNN), which extracts short-term temporal patterns from the input sequence (Eq.~\ref{eq:cnn}). This captures local trends and periodic fluctuations in the PM$_{2.5}$ signal.The output of the CNN is then passed through stacked GRU (Gated Recurrent Unit) layers to model long-term dependencies across timesteps (Eq.~\ref{eq:gru}). This sequence modelling step ensures that both immediate and delayed temporal correlations are captured effectively. Next, a Multi-Layer Perceptron (MLP) processes the GRU output to produce temporally encoded features for each forecasting timestep (Eq.~\ref{eq:mlp}). These temporal features are concatenated with spatiotemporal features obtained from the graph-based module to form fused representations (Eq.~\ref{eq:fusion}). 
\subsubsection{Low-Rank Adaptation Module}
\vspace{1em}
\noindent To refine this combined representation, the model uses a \textbf{Linear Attention} mechanism, which replaces expensive dot-product attention with an efficient kernel-based approximation (Eqs.~\ref{eq:kernel},~\ref{eq:linear_attn}).The attention projection matrices for query, key, and value are enhanced with \textbf{LoRA (Low-Rank Adaptation)} (Eq.~\ref{eq:lora}), enabling efficient parameter adaptation during training. The refined features produced by attention are then passed through a final MLP to generate the forecasted PM$_{2.5}$ values over the prediction horizon (Eq.~\ref{eq:final_pred}).
\begin{equation}
	h_{\text{conv}} = f_{\text{conv}}(x_t^{\text{PM}_{2.5}})
	\label{eq:cnn}
\end{equation}
\begin{equation}
	\tilde{h}_{t_z}^{\text{out}} = f'_{\text{gru}}(\dots f'_{\text{gru}}(h_{\text{conv}})), \quad t_z \in (t+1, \dots, t+u)
	\label{eq:gru}
\end{equation}
\begin{equation}
	\{h_{t_z}^{\text{out}}\}_{t_z} = \text{MLP}(\{\tilde{h}_{t_z}^{\text{out}}\}_{t_z})
	\label{eq:mlp}
\end{equation}
\begin{equation}
	h_{t_z}^{\text{fusion}} = h_{t_z}^{\text{out}} \oplus h_{t_z}
	\label{eq:fusion}
\end{equation}
\begin{equation}
	\phi(x) = \text{ELU}(x) + 1
	\label{eq:kernel}
\end{equation}
\begin{equation}
	\text{Attention}(Q, K, V) \approx \phi(Q) \cdot \left( \phi(K)^T V \right)
	\label{eq:linear_attn}
\end{equation}
\begin{equation}
	W_Q = W_Q^0 + A_Q B_Q, \quad W_K = W_K^0 + A_K B_K, \quad W_V = W_V^0 + A_V B_V
	\label{eq:lora}
\end{equation}
\begin{equation}
	\hat{h}_{t_z} = \phi(Q) \cdot \left( \phi(K)^T V \right)
\end{equation}
\begin{equation}
	(\hat{x}_t^{\text{PM}_{2.5}}, \dots, \hat{x}_{t+u}^{\text{PM}_{2.5}}) = \text{MLP}_{\text{out}}(\{\hat{h}_{t_z}\})
	\label{eq:final_pred}
\end{equation}
To train the model effectively while ensuring generalization, a loss function combining the Mean Squared Error (MSE) with L2 regularization is adopted. The MSE term penalizes the difference between predicted and actual PM$_{2.5}$ values, thereby promoting accurate forecasting. Simultaneously, the L2 regularization term constrains model parameters by penalizing large weights, which helps reduce the risk of overfitting. The overall objective function is defined as:
\begin{equation}
	\mathcal{L}(\theta) = \frac{1}{N} \sum_{i=1}^{N} \left( y_i - \hat{y}_i \right)^2 + \lambda \|\theta\|_2^2
	\label{eq:loss}
\end{equation}
where \( y_i \) and \( \hat{y}_i \) are the ground truth and predicted values respectively, \( \theta \) denotes the trainable parameters of the model, \( \|\theta\|_2^2 \) is the L2 norm of the parameters, and \( \lambda \) is the regularization coefficient that balances accuracy and complexity.

\subsection{Evaluation Metrics}
The predictive performance is evaluated using error-based and event-based metrics, summarized in Table~\ref{tab:metrics}.
\vspace{-2em}
\begin{table}[H]
	\small
	\centering
	\caption{Compact summary of evaluation metrics with formulas and brief descriptions.}
	\label{tab:metrics}
	\renewcommand{\arraystretch}{1.3}
	\begin{tabular}{llp{7cm}}
		\hline
		\textbf{Metric} & \textbf{Formula} & \textbf{Description} \\
		\hline
		RMSE & $\displaystyle \sqrt{\frac{1}{n} \sum_{i=1}^{n}(y_i - \hat{y}_i)^2}$ & Average magnitude of squared errors. Lower values indicate better accuracy. \\
		\hline
		MAE & $\displaystyle \frac{1}{n} \sum_{i=1}^{n} |y_i - \hat{y}_i|$ & Average absolute deviation between predictions and actual values. \\
		\hline
		MAPE & $\displaystyle \frac{100}{n} \sum_{i=1}^{n} \left| \frac{y_i - \hat{y}_i}{y_i} \right|$ & Average error as a percentage of actual values; lower is better. \\
		\hline
		$R^2$ & $\displaystyle 1 - \frac{\sum_{i=1}^{n} (y_i - \hat{y}_i)^2}{\sum_{i=1}^{n} (y_i - \bar{y})^2}$ & Proportion of variance explained by the model; closer to 1 is better. \\
		\hline
		CSI & $\displaystyle \frac{\text{Hits}}{\text{Hits} + \text{Misses} + \text{False Alarms}}$ & Evaluates event prediction accuracy considering hits, misses, and false alarms. Higher is better. \\
		\hline
		POD & $\displaystyle \frac{\text{Hits}}{\text{Hits} + \text{Misses}}$ & Fraction of actual exceedances correctly predicted; closer to 1 indicates high sensitivity. \\
		\hline
		FAR & $\displaystyle \frac{\text{False Alarms}}{\text{Hits} + \text{False Alarms}}$ & Proportion of false alerts; lower indicates more reliable forecasting. \\
		\hline
	\end{tabular}
\end{table}

\section{Results}
This section presents the qualitative and quantitative evaluation of the proposed model. The performance of the model is assessed using multiple metrics and visual analyses to demonstrate its effectiveness in capturing spatiotemporal variations in PM\textsubscript{2.5} concentrations.

\subsection{Quantitative Result} \label{sec:result1}
Table~\ref{tab:model_performace} presents a comprehensive evaluation of various deep learning models for PM$_{2.5}$ prediction using multiple performance metrics, including MAE, RMSE, $R^{2}$, MAPE, CSI, POD, and FAR. The comparison includes traditional sequence models like ANN, RNN, LSTM, and GRU as well as several architectural variants of the proposed Spatio-Temporal Graph Neural Network (SA-GNN LoRA), each incorporating enhancements such as Graph Attention (GAT) layers, multi-head self-attention, adapter layers, and Linear Attention with Low-Rank Adaptation (LoRA).
\begin{table}[H]
	\centering
	\caption{Performance Comparison of Models for sample type (r=20, u=8) on K-Fold training technique}
	\label{tab:model_performace}
	
	\begin{tabular}{p{4.2cm}ccccccc}
		\hline
		\textbf{Model} & \textbf{MAE} & \textbf{RMSE} & \textbf{R\textsuperscript{2}} & \textbf{MAPE (\%)} & \textbf{CSI} & \textbf{POD} & \textbf{FAR} \\
		\hline
		ANN & 4.70 & 18.40 & 0.77 & 4.78 & 90.24 & 90.67 & 0.51 \\
		\hline
		LSTM & 3.75 & 15.85 & 0.83 & 5.59 & 89.62 & 99.89 & 10.29 \\
		\hline
		GRU & 11.72 & 19.01 & 0.76 & 96.39 & 87.34 & 90.09 & 2.75 \\
		\hline
		RNN & 10.56 & 18.43 & 0.87 & 7.89 & 93.36 & 93.61 & 0.28 \\
		\hline
		GNN & 5.88 & 8.08 & 0.90 & 6.19 & 91.76 & 92.64 & 5.24 \\
		\hline
		\textbf{SA-GNN-LoRA} & 4.19 & 6.84 & 0.95 & 5.76 & 95.79 & 98.88 & 3.15 \\
		
		\hline
	\end{tabular}
\end{table}

Among the baseline models, sequence-based architectures (RNN, LSTM, GRU) outperform the feedforward ANN due to their ability to capture temporal dependencies in air quality time series. Notably, the LSTM model achieves the lowest MAE (3.75), while the RNN model records the highest $R^{2}$ (0.87), indicating strong temporal modelling capabilities. GRU delivers comparable results; however, its slightly higher error values suggest sensitivity to temporal irregularities frequently observed in environmental datasets.

The SA-GNN variants consistently outperform the baseline models, emphasizing the importance of explicitly modelling spatial dependencies through graph-based structures. The most effective configuration SA-GNN with Linear Attention and LoRA achieves the highest $R^{2}$ (0.95), the lowest RMSE (6.84), and excellent detection performance (CSI: 95.79, POD: 98.88), while maintaining a low FAR (3.15). These results highlight the advantages of integrating efficient attention mechanisms with parameter-efficient tuning strategies such as LoRA, which enhance model performance without incurring significant computational overhead.

\subsubsection{Ablation Study}
The ablation study reveals that removing the Linear Attention component significantly degrades performance, particularly in variants that rely solely on GAT layers(Table~\ref{tab:model_performance}). This outcome confirms that capturing temporal dependencies remains critical, even when spatial graph structures are employed, thereby underscoring the complementary roles of spatial and temporal attention mechanisms. Furthermore, although architectures incorporating Double GAT and Self-Head Attention achieve high CSI and POD scores, they exhibit elevated FAR values (7.43 and 4.39, respectively), indicating a propensity to over-predict pollution events, which may reduce their reliability in real-world alert systems.
\begin{table}[ht]
	\centering
	\caption{Performance Comparison of Models for sample type (r=20, u=8) on K-Fold training technique}
	\label{tab:model_performance}
	\begin{tabular}{p{4.2cm}ccccccc}
		\hline
		\textbf{Model} & \textbf{MAE} & \textbf{RMSE} & \textbf{R\textsuperscript{2}} & \textbf{MAPE (\%)} & \textbf{CSI} & \textbf{POD} & \textbf{FAR} \\
		\hline
		SA-GNN (Single GAT Layer, w/o LinearAttention) & 8.56 & 18.11 & 0.82 & 11.45 & 88.42 & 95.81 & 7.02 \\
		\hline
		SA-GNN (Double GAT Layer, w/o LinearAttention) & 9.95 & 16.82 & 0.88 & 11.50 & 89.95 & 96.96 & 7.43 \\
		\hline
		SA-GNN (Self-Head Attention) & 10.22 & 17.64 & 0.86 & 11.02 & 92.31 & 96.40 & 4.39 \\
		\hline
		SA-GNN (Multi-Head Attention) & 5.39 & 9.72 & 0.92 & 6.26 & 81.71 & 82.95 & 1.79 \\
		\hline
		SA-GNN (With Adapter Layer) & 3.88 & 8.97 & 0.91 & 4.64 & 92.77 & 95.03 & 2.49 \\
		\hline
		SA-GNN (Linear Attention) & 5.70 & 8.89 & 0.93 & 6.63 & 78.95 & 79.50 & 0.85 \\
		\hline
		SA-GNN (Linear Attention + LoRA) & 4.19 & 6.84 & 0.95 & 5.76 & 95.79 & 98.88 & 3.15 \\
		\hline
	\end{tabular}
\end{table}

Other enhancements within the SA-GNN framework also demonstrate strong performance. The Multi-Head Attention variant with 12 attention heads achieves a high $R^2$ value (0.92) and provides a more balanced trade-off between detection accuracy and false alarm rate compared to variants with fewer heads. Likewise, the Adapter Layer variant performs consistently well across all evaluation metrics and records the lowest MAPE (4.64), suggesting that lightweight architectural modules can effectively improve feature interaction and model generalization.

\subsubsection{Effect of Meteorological Factors}
In the model study, omitting all meteorological and land-use variables from the SA-GNN architecture resulted in a marked decline in forecasting performance. The coefficient of determination ($R^2$) decreased to 0.87, while the Mean Absolute Error (MAE) increased to approximately $7.20\ \mu\text{g}/\text{m}^3$ and the Root Mean Square Error (RMSE) rose to about $12.0\ \mu\text{g}/\text{m}^3$. To identify the most influential external features, both Local Interpretable Model-Agnostic Explanations (LIME) and Layer-wise Relevance Propagation (LRP) were applied to the full set of ten selected input variables.

The analysis revealed that the feature corresponding to bike speed consistently exhibited negligible or negative importance across both interpretability techniques and was therefore excluded from subsequent experiments. In contrast, features such as wind direction, wind speed, temperature, commercial-zone density, and residential-zone density demonstrated consistently positive relevance scores, indicating their significant contributions to the model’s predictive capability.

Interestingly, although PM${2.5}$ values are the primary target variable, their local importance scores in LIME were relatively low. This can be attributed to the normalization of features and the presence of strongly correlated external variables. Since PM${2.5}$ values were rescaled during preprocessing, their numeric magnitude became comparable to other features, which reduces the apparent effect of small perturbations in LIME’s local explanation. Additionally, meteorological and land-use variables capture much of the variance in PM${2.5}$ across nodes and time steps, meaning the model can rely on these correlated features for predictions, further reducing the measured local importance of PM${2.5}$ itself.
\begin{figure}[ht]
	\centering
	\includegraphics[width=0.8\linewidth, height=4cm]{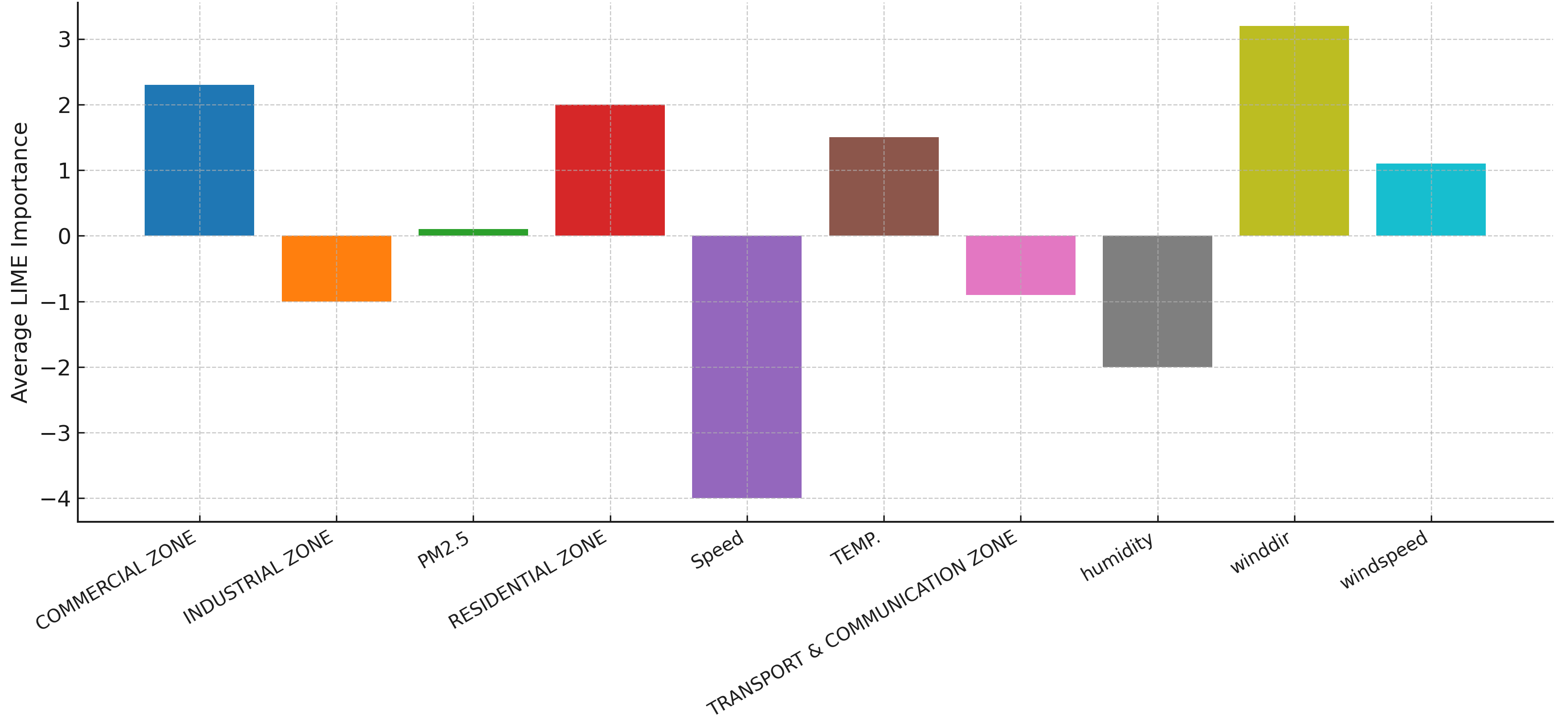}\
	\caption{LIME Feature Importance plot of SA-GNN LoRA model}
	\label{fig:lime}
\end{figure}
The LIME based feature importance analysis (Fig.~\ref{fig:lime}) confirmed that meteorological and land-use variables made consistently positive contributions across all nodes and time steps. Integrating these five key external factors into the SA-GNN model, alongside traffic and temporal features, significantly enhanced performance raising the $R^2$ score from 0.87 to 0.95 and reducing both MAE and RMSE. These results demonstrate the strong influence of contextual features and validate the proposed SA-GNN architecture with Linear Attention and LoRA as an accurate, efficient, and deployable framework for spatiotemporal air quality forecasting.
\subsection{Qualitative Results}\label{sec:result2}
\begin{figure}[H]
	\begin{minipage}[t]{0.45\linewidth}
		\centering
		\includegraphics[height=3.5cm]{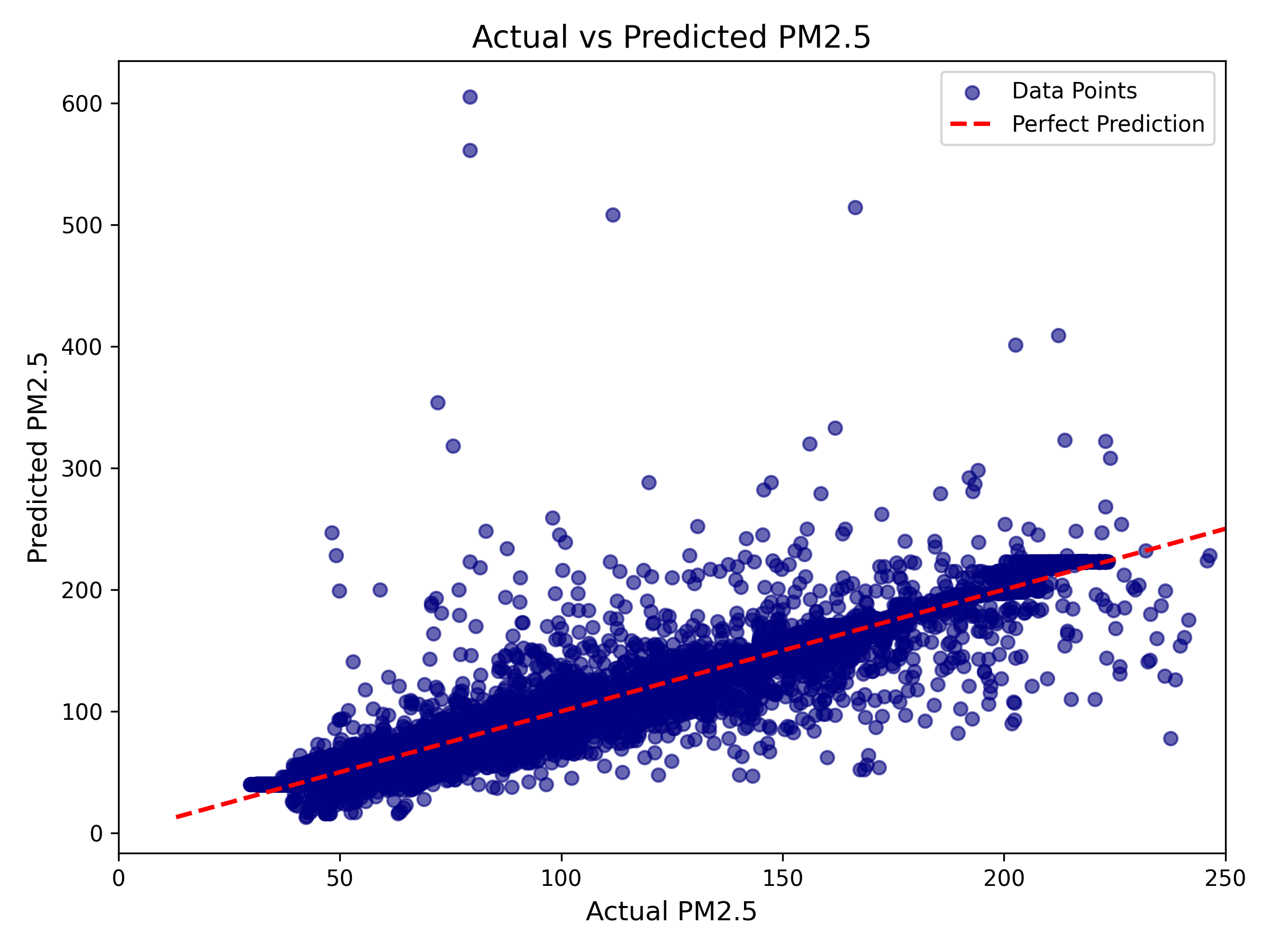}
		\caption*{\textbf{(A)}}
	\end{minipage}
	\begin{minipage}[t]{0.45\linewidth}
		\centering
		\includegraphics[height=3.5cm]{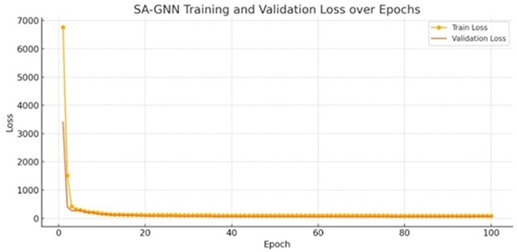}
		\caption*{\textbf{(B)}}
	\end{minipage}
	\caption{\textbf{(A)} Actual vs. Predicted PM\textsubscript{2.5} concentrations; \textbf{(B)} SA-GNN Training and validation loss across epochs}
	\label{fig:scatter_plot}
\end{figure}

\begin{figure}[H]
	\centering
	
	\begin{subfigure}[b]{0.48\linewidth}
		\centering
		\includegraphics[width=\linewidth, height=4.5cm]{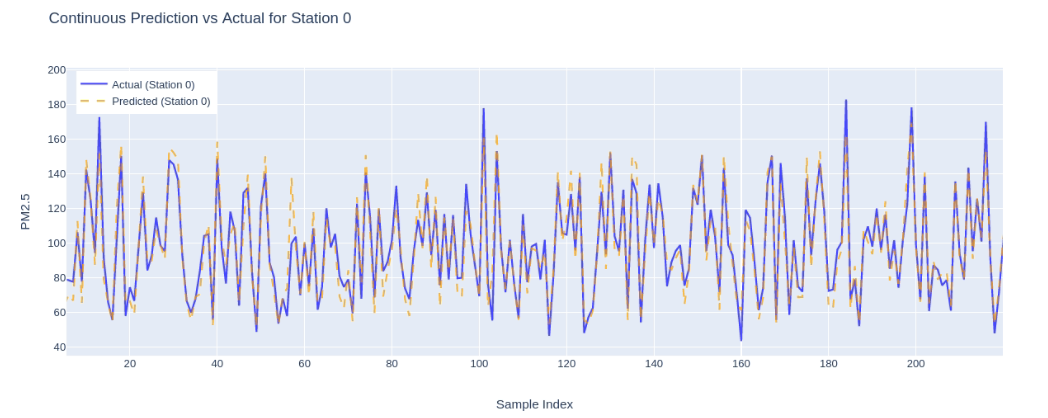}
		\caption*{\textbf{(A)}}
	\end{subfigure}
	\hfill
	\begin{subfigure}[b]{0.48\linewidth}
		\centering
		\includegraphics[width=\linewidth, height=4cm]{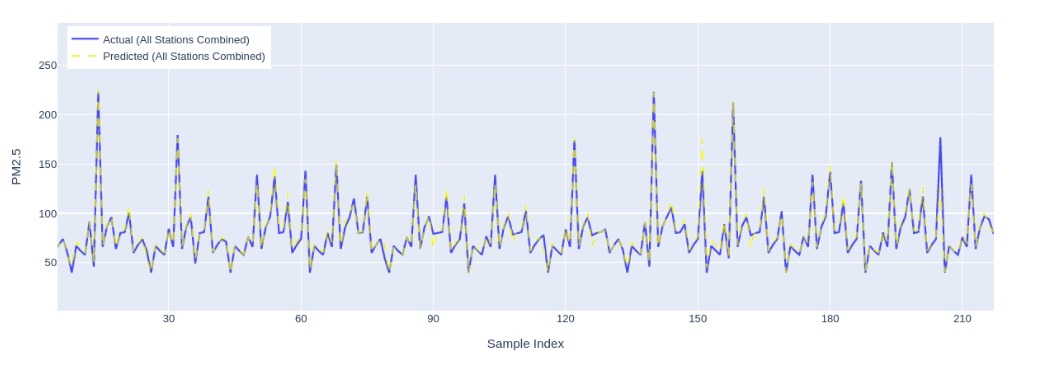}
		\caption*{\textbf{(B)}}
	\end{subfigure}
	
	\vspace{0.6em}
	
	\begin{subfigure}[b]{0.85\linewidth}
		\centering
		\includegraphics[width=\linewidth, height=4cm]{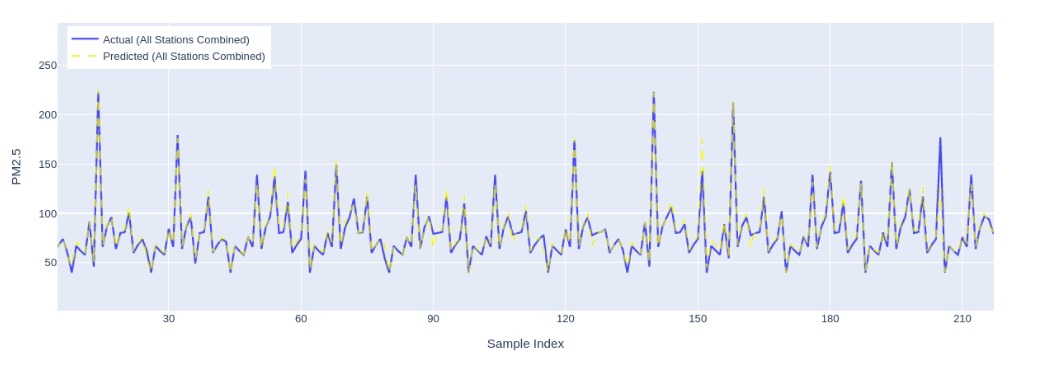}
		\caption*{\textbf{(C)}}
	\end{subfigure}
	
	\caption{\textbf{(A)} Prediction at Station 0; \textbf{(B)} Prediction across all stations; \textbf{(C)} Prediction at Station 7.}
	\label{fig:predicted_station}
\end{figure}

\noindent\textbf{Real-World Applicability of the Proposed Model}
The practical utility of the proposed model for \textbf{real-time air quality forecasting} is demonstrated through five key visualizations. These collectively confirm the model’s suitability for deployed environmental monitoring systems and on-device predictions to support pollution control interventions.
Fig.~\ref{fig:scatter_plot}A presents a scatter plot of actual versus predicted PM$_{2.5}$ concentrations. The clustering of data points along the diagonal line ($y = x$) indicates high predictive accuracy, suggesting that the model effectively replicates observed pollutant concentrations across a broad range of values. While minor deviations appear at higher concentration levels, their limited magnitude does not substantially impact practical deployment, particularly for high-pollution alerts.
Fig. 6B displays the training and validation loss trends across epochs. The tight convergence between the two curves reflects robust generalization, a key requirement for real-time deployment in edge-based or embedded systems.
Fig.~\ref{fig:scatter_plot}B illustrate the model training loss curve, respectively. Fig.~\ref{fig:predicted_station}A demonstrates the model’s ability to accurately capture short-term station-level fluctuations, including sharp pollutant spikes at a specific monitoring station. Fig.~\ref{fig:predicted_station}B validates the model’s effectiveness across all stations, highlighting its robust network-wide forecasting capability. Furthermore, Fig.~\ref{fig:predicted_station}C shows that the model remains resilient under complex temporal dynamics, effectively capturing rapid pollutant variations with high fidelity at Station 7. Therefore, the SA-GNN LoRA model is not only effective from a research standpoint but also highly suitable for \textbf{real-world PM$_{2.5}$ forecasting applications}, offering tangible benefits for public health, environmental policy, and urban sustainability.

\section{Conclusion}\label{sec:conclusion}
\noindent Accurate short-term forecasting and real-time monitoring of PM$_{2.5}$ concentrations are essential for safeguarding public health, particularly in rapidly urbanizing cities. This study introduces SA-GNN-LoRA, a spatiotemporal attention-based graph neural network tailored for PM$_{2.5}$ prediction using mobile sensing data collected along a 14\,km urban route in Surat, India. The dataset comprises 53 mobile sensing runs, capturing detailed spatial and temporal variations in pollutants. The proposed model captures spatial dependencies using two Graph Attention Network (GAT) and temporal dependencies through a hybrid 1D-CNN and GRU module. A cluster-based spatial grouping strategy is introduced, where monitoring points with similar pollution dynamics are assigned to common clusters, each modelled by a dedicated GRU encoder. This design allows for localized temporal modelling and enhances the model’s ability to reflect spatial heterogeneity.
The model is trained using 20 historical time steps (60 seconds) to predict 8 future steps (24 seconds). Empirical evaluations show that SA-GNN-LoRA significantly outperforms traditional baselines including LSTM, GRU, ANN, and RNN. It achieves an RMSE of $6.84\,\mu\text{g}/\text{m}^3$, MAE of $4.19\,\mu\text{g}/\text{m}^3$, and an $R^2$ score of 0.95, indicating strong predictive performance and generalization. Overall, the study holds a good promise demonstrating the use of hybrid deep learning models in predicting spatiotemporal PM2.5 patterns captured using mobile devices for the purpose of monitoring and assessment of urban air quality parameters along the arterial corridor. 

\vspace{1em}

\noindent\textbf{AUTHOR CONTRIBUTIONS} \\[0.5em]
The authors confirm contribution to the paper as follows: \textbf{Study conception and design}: All authors; \textbf{Data collection}: Om Chiddarwar and Priyanka Mandal; \textbf{Analysis and interpretation of results}: All authors; \textbf{Draft manuscript preparation}: All authors. All authors reviewed the results and approved the final version of the manuscript.

\vspace{1em}

\noindent\textbf{FUNDING} \\[0.5em]
This research was supported by sponsoring agencies, Ministry of Housing \& Urban Affairs (4/254) and GUJCOST (4/316).

\vspace{1em}

\noindent\textbf{ACKNOWLEDGEMENT} \\[0.5em]
The authors acknowledge the Department of AI and Computer Science, SVNIT Surat, for providing computational resources on the Dual NVIDIA H100 NVL-based GPU cluster, which enabled large-scale model training and experimentation.



\begin{thebibliography}{34}
	\expandafter\ifx\csname natexlab\endcsname\relax\def\natexlab#1{#1}\fi
	\providecommand{\url}[1]{\texttt{#1}}
	\providecommand{\href}[2]{#2}
	\providecommand{\path}[1]{#1}
	\providecommand{\DOIprefix}{doi:}
	\providecommand{\ArXivprefix}{arXiv:}
	\providecommand{\URLprefix}{URL: }
	\providecommand{\Pubmedprefix}{pmid:}
	\providecommand{\doi}[1]{\href{http://dx.doi.org/#1}{\path{#1}}}
	\providecommand{\Pubmed}[1]{\href{pmid:#1}{\path{#1}}}
	\providecommand{\bibinfo}[2]{#2}
	\ifx\xfnm\relax \def\xfnm[#1]{\unskip,\space#1}\fi
	\bibitem[{All{\'e}on et~al.(2020)All{\'e}on, Jauvion, Quennehen and
		Lissmyr}]{alleon2020plumenet}
	\bibinfo{author}{All{\'e}on, A.}, \bibinfo{author}{Jauvion, G.},
	\bibinfo{author}{Quennehen, B.}, \bibinfo{author}{Lissmyr, D.},
	\bibinfo{year}{2020}.
	\newblock \bibinfo{title}{Plumenet: Large-scale air quality forecasting using a
		convolutional lstm network}.
	\newblock \bibinfo{journal}{arXiv preprint arXiv:2006.09204}
	\DOIprefix\doi{10.48550/arXiv.2006.09204}.
	\bibitem[{Apte et~al.(2017)Apte, Messier, Gani, Brauer, Kirchstetter, Lunden,
		Marshall, Portier, Vermeulen and Hamburg}]{apte2017high}
	\bibinfo{author}{Apte, J.S.}, \bibinfo{author}{Messier, K.P.},
	\bibinfo{author}{Gani, S.}, \bibinfo{author}{Brauer, M.},
	\bibinfo{author}{Kirchstetter, T.W.}, \bibinfo{author}{Lunden, M.M.},
	\bibinfo{author}{Marshall, J.D.}, \bibinfo{author}{Portier, C.J.},
	\bibinfo{author}{Vermeulen, R.C.H.}, \bibinfo{author}{Hamburg, S.P.},
	\bibinfo{year}{2017}.
	\newblock \bibinfo{title}{High‑resolution air pollution mapping with google
		street view cars: Exploiting big data}.
	\newblock \bibinfo{journal}{Environmental Science \& Technology}
	\bibinfo{volume}{51}, \bibinfo{pages}{6999--7008}.
	\newblock \DOIprefix\doi{10.1021/acs.est.7b00891}.
	\bibitem[{Author(2021)}]{theoryguided2021}
	\bibinfo{author}{Author, A.}, \bibinfo{year}{2021}.
	\newblock \bibinfo{title}{A theory‑guided graph networks based pm2.5
		forecasting method}.
	\newblock \bibinfo{journal}{Science of The Total Environment} \URLprefix
	\url{https://doi.org/10.1016/j.envpol.2021.118569}.
	\bibitem[{Balakrishnan et~al.(2019)Balakrishnan, Dey, Gupta and et.
		al}]{balakrishnan2019impact}
	\bibinfo{author}{Balakrishnan, K.}, \bibinfo{author}{Dey, S.},
	\bibinfo{author}{Gupta, T.}, \bibinfo{author}{et. al, D.},
	\bibinfo{year}{2019}.
	\newblock \bibinfo{title}{The impact of air pollution on deaths, disease
		burden, and life expectancy across the states of india: the global burden of
		disease study 2017}.
	\newblock \bibinfo{journal}{The Lancet Planetary Health} \bibinfo{volume}{3},
	\bibinfo{pages}{e26--e39}.
	\newblock \DOIprefix\doi{10.1016/S2542-5196(18)30261-4}.
	\bibitem[{Castell et~al.(2017)Castell, Dauge, Schneider, Vogt, Lerner, Fishbain
		and Bartonova}]{castell2017can}
	\bibinfo{author}{Castell, N.}, \bibinfo{author}{Dauge, F.R.},
	\bibinfo{author}{Schneider, P.}, \bibinfo{author}{Vogt, M.},
	\bibinfo{author}{Lerner, U.}, \bibinfo{author}{Fishbain, B.},
	\bibinfo{author}{Bartonova, A.}, \bibinfo{year}{2017}.
	\newblock \bibinfo{title}{Can commercial low‑cost sensor platforms contribute
		to air quality monitoring and exposure estimates?}
	\newblock \bibinfo{journal}{Environment International} \bibinfo{volume}{99},
	\bibinfo{pages}{293--302}.
	\newblock \DOIprefix\doi{10.1016/j.envint.2016.12.007}.
	\bibitem[{Chen et~al.(2022)Chen, Zhang and Li}]{chen2022hybrid}
	\bibinfo{author}{Chen, H.}, \bibinfo{author}{Zhang, L.}, \bibinfo{author}{Li,
		W.}, \bibinfo{year}{2022}.
	\newblock \bibinfo{title}{A hybrid spatiotemporal model for short-term air
		quality forecasting using cnn and bigru}.
	\newblock \bibinfo{journal}{Ecological Indicators} \bibinfo{volume}{139},
	\bibinfo{pages}{108913}.
	\bibitem[{Cohen et~al.(2017)Cohen, Brauer, Burnett, Anderson, Frostad, Estep
		and Forouzanfar}]{cohen2017estimates}
	\bibinfo{author}{Cohen, A.J.}, \bibinfo{author}{Brauer, M.},
	\bibinfo{author}{Burnett, R.}, \bibinfo{author}{Anderson, H.R.},
	\bibinfo{author}{Frostad, J.}, \bibinfo{author}{Estep, K.},
	\bibinfo{author}{Forouzanfar, M.H.}, \bibinfo{year}{2017}.
	\newblock \bibinfo{title}{Estimates and 25-year trends of the global burden of
		disease attributable to ambient air pollution}.
	\newblock \bibinfo{journal}{The Lancet} \bibinfo{volume}{389},
	\bibinfo{pages}{1907--1918}.
	\newblock \DOIprefix\doi{10.1016/S0140-6736(17)30505-6}.
	\bibitem[{Cui et~al.(2023)Cui, Zhang and Li}]{cui2023comparative}
	\bibinfo{author}{Cui, X.}, \bibinfo{author}{Zhang, W.}, \bibinfo{author}{Li,
		M.}, \bibinfo{year}{2023}.
	\newblock \bibinfo{title}{Comparative analysis of time series forecasting
		models for urban air pollution prediction}.
	\newblock \bibinfo{journal}{Environmental Modelling \& Software}
	\bibinfo{volume}{165}, \bibinfo{pages}{105662}.
	\newblock \DOIprefix\doi{https://doi.org/10.1016/j.aej.2024.03.031}.
	\bibitem[{Gao et~al.(2020)Gao, Wang, Liu and Zhu}]{gao2020predicting}
	\bibinfo{author}{Gao, J.}, \bibinfo{author}{Wang, L.}, \bibinfo{author}{Liu,
		Y.}, \bibinfo{author}{Zhu, W.}, \bibinfo{year}{2020}.
	\newblock \bibinfo{title}{Predicting pm2.5 concentration using xgboost based on
		spatiotemporal characteristics}.
	\newblock \bibinfo{journal}{Science of The Total Environment}
	\bibinfo{volume}{724}, \bibinfo{pages}{138201}.
	\newblock \DOIprefix\doi{https://doi.org/10.3390/su132112071}.
	\bibitem[{Hammer et~al.(2020)Hammer, van Donkelaar, Li, Lyapustin, Sayer, Hsu
		and Martin}]{hammer2020global}
	\bibinfo{author}{Hammer, M.S.}, \bibinfo{author}{van Donkelaar, A.},
	\bibinfo{author}{Li, C.}, \bibinfo{author}{Lyapustin, A.},
	\bibinfo{author}{Sayer, A.M.}, \bibinfo{author}{Hsu, N.C.},
	\bibinfo{author}{Martin, R.V.}, \bibinfo{year}{2020}.
	\newblock \bibinfo{title}{Global estimates and long-term trends of fine
		particulate matter concentrations (1998–2018)}.
	\newblock \bibinfo{journal}{Environmental Science \& Technology}
	\bibinfo{volume}{54}, \bibinfo{pages}{7879--7890}.
	\newblock \DOIprefix\doi{https://doi.org/10.1021/acs.est.0c01764}.
	\bibitem[{Hettige et~al.(2024)Hettige, Ji et~al.}]{hettige2024airphynet}
	\bibinfo{author}{Hettige, K.H.}, \bibinfo{author}{Ji, J.}, et~al.,
	\bibinfo{year}{2024}.
	\newblock \bibinfo{title}{Airphynet: Harnessing physics‑guided neural
		networks for air quality prediction}, in: \bibinfo{booktitle}{ICLR 2024}.
	\newblock \DOIprefix\doi{10.48550/arXiv.2402.03784}.
	\bibitem[{Hu et~al.(2022)Hu, Shen, Wallis, Allen-Zhu, Li, Wang and
		Chen}]{hu2022lora}
	\bibinfo{author}{Hu, E.J.}, \bibinfo{author}{Shen, Y.},
	\bibinfo{author}{Wallis, P.}, \bibinfo{author}{Allen-Zhu, Z.},
	\bibinfo{author}{Li, Y.}, \bibinfo{author}{Wang, W.}, \bibinfo{author}{Chen,
		Y.}, \bibinfo{year}{2022}.
	\newblock \bibinfo{title}{Lora: Low-rank adaptation of large language models}.
	\newblock \bibinfo{journal}{arXiv preprint arXiv:2106.09685}
	\DOIprefix\doi{https://doi.org/10.1145/3676151.3719377}.
	\bibitem[{Kumar and Goyal(2011)}]{kumar2011arima}
	\bibinfo{author}{Kumar, A.}, \bibinfo{author}{Goyal, P.}, \bibinfo{year}{2011}.
	\newblock \bibinfo{title}{Forecasting of air quality in delhi using principal
		component regression technique}.
	\newblock \bibinfo{journal}{Atmospheric Pollution Research}
	\bibinfo{volume}{2}, \bibinfo{pages}{436--444}.
	\newblock \DOIprefix\doi{https://doi.org/10.5094/APR.2011.050}.
	\bibitem[{Kumar et~al.(2015)Kumar, Morawska, Martani, Biskos, Neophytou,
		Di~Sabatino and Britter}]{kumar2015rise}
	\bibinfo{author}{Kumar, P.}, \bibinfo{author}{Morawska, L.},
	\bibinfo{author}{Martani, C.}, \bibinfo{author}{Biskos, G.},
	\bibinfo{author}{Neophytou, M.}, \bibinfo{author}{Di~Sabatino, S.},
	\bibinfo{author}{Britter, R.}, \bibinfo{year}{2015}.
	\newblock \bibinfo{title}{The rise of low-cost sensing for managing air
		pollution in cities}.
	\newblock \bibinfo{journal}{Environment International} \bibinfo{volume}{75},
	\bibinfo{pages}{199--205}.
	\newblock \DOIprefix\doi{10.1016/j.envint.2014.11.019}.
	\bibitem[{Li et~al.(2017a)Li, Peng, Yao, Cui, Hu, You and Chi}]{li2017lstm}
	\bibinfo{author}{Li, X.}, \bibinfo{author}{Peng, L.}, \bibinfo{author}{Yao,
		X.}, \bibinfo{author}{Cui, S.}, \bibinfo{author}{Hu, Y.},
	\bibinfo{author}{You, C.}, \bibinfo{author}{Chi, T.}, \bibinfo{year}{2017}a.
	\newblock \bibinfo{title}{Long short-term memory neural network for air
		pollutant concentration predictions: Method development and evaluation}.
	\newblock \bibinfo{journal}{Environmental Pollution} \bibinfo{volume}{231},
	\bibinfo{pages}{997--1004}.
	\newblock \DOIprefix\doi{https://doi.org/10.1016/j.envpol.2017.08.114}.
	\bibitem[{Li et~al.(2017b)Li, Yu, Shahabi and Liu}]{li2017diffusion}
	\bibinfo{author}{Li, Y.}, \bibinfo{author}{Yu, R.}, \bibinfo{author}{Shahabi,
		C.}, \bibinfo{author}{Liu, Y.}, \bibinfo{year}{2017}b.
	\newblock \bibinfo{title}{Diffusion convolutional recurrent neural network:
		Data-driven traffic forecasting}, in: \bibinfo{booktitle}{Proceedings of the
		International Conference on Learning Representations}.
	\bibitem[{Liang et~al.(2015)Liang, Zou, Guo, Li, Zhang, Zhang and
		Chen}]{liang2015assessing}
	\bibinfo{author}{Liang, X.}, \bibinfo{author}{Zou, T.}, \bibinfo{author}{Guo,
		B.}, \bibinfo{author}{Li, S.}, \bibinfo{author}{Zhang, H.},
	\bibinfo{author}{Zhang, S.}, \bibinfo{author}{Chen, S.},
	\bibinfo{year}{2015}.
	\newblock \bibinfo{title}{Assessing beijing's pm2.5 pollution: severity,
		weather impact, apec and winter heating}.
	\newblock \bibinfo{journal}{Proceedings of the Royal Society A}
	\bibinfo{volume}{471}, \bibinfo{pages}{20150257}.
	\newblock \DOIprefix\doi{https://doi.org/10.1098/rspa.2015.0257}.
	\bibitem[{Ma et~al.(2020)Ma, Ma, Zhang and Wang}]{ma2020learning}
	\bibinfo{author}{Ma, X.}, \bibinfo{author}{Ma, L.}, \bibinfo{author}{Zhang,
		F.}, \bibinfo{author}{Wang, Y.}, \bibinfo{year}{2020}.
	\newblock \bibinfo{title}{Learning traffic as images: a deep convolutional
		neural network for large-scale transportation network speed prediction}.
	\newblock \bibinfo{journal}{Sensors} \bibinfo{volume}{20},
	\bibinfo{pages}{2554}.
	\bibitem[{Pope~III and Dockery(2006)}]{pope2006health}
	\bibinfo{author}{Pope~III, C.A.}, \bibinfo{author}{Dockery, D.W.},
	\bibinfo{year}{2006}.
	\newblock \bibinfo{title}{Health effects of fine particulate air pollution:
		lines that connect}.
	\newblock \bibinfo{journal}{Journal of the Air \& Waste Management Association}
	\bibinfo{volume}{56}, \bibinfo{pages}{709--742}.
	\newblock \DOIprefix\doi{10.1080/10473289.2006.10464485}.
	\bibitem[{Qi et~al.(2025)}]{qi2019gclstm}
	\bibinfo{author}{Qi, X.}, et~al., \bibinfo{year}{2025}.
	\newblock \bibinfo{title}{Hybrid graph convolutional lstm model for
		spatio-temporal air quality forecasting}.
	\newblock \URLprefix \url{https://doi.org/10.1007/s11869-025-01713-8}.
	\bibinfo{note}{transfer learning with GC-LSTM}.
	\bibitem[{Samal et~al.(2024)Samal, Roy and Panda}]{samal2024autotcn}
	\bibinfo{author}{Samal, S.}, \bibinfo{author}{Roy, A.}, \bibinfo{author}{Panda,
		R.}, \bibinfo{year}{2024}.
	\newblock \bibinfo{title}{Auto-tcn: Automated temporal convolutional network
		architecture search for time series forecasting}, in:
	\bibinfo{booktitle}{Proceedings of the AAAI Conference on Artificial
		Intelligence}, pp. \bibinfo{pages}{5431--5439}.
	\bibitem[{Seo and Liu(2019)}]{seo2019differentiable}
	\bibinfo{author}{Seo, S.}, \bibinfo{author}{Liu, Y.}, \bibinfo{year}{2019}.
	\newblock \bibinfo{title}{Differentiable physics-informed graph networks}, in:
	\bibinfo{booktitle}{arXiv}.
	\newblock \URLprefix \url{https://arxiv.org/abs/1902.02950}.
	\bibitem[{Shaddick et~al.(2018)Shaddick, Thomas, Amini, Broday, Cohen, Frostad,
		Green, Gumy, Liu, Martin et~al.}]{shaddick2018data}
	\bibinfo{author}{Shaddick, G.}, \bibinfo{author}{Thomas, M.L.},
	\bibinfo{author}{Amini, H.}, \bibinfo{author}{Broday, D.},
	\bibinfo{author}{Cohen, A.}, \bibinfo{author}{Frostad, J.},
	\bibinfo{author}{Green, A.}, \bibinfo{author}{Gumy, S.},
	\bibinfo{author}{Liu, Y.}, \bibinfo{author}{Martin, R.V.}, et~al.,
	\bibinfo{year}{2018}.
	\newblock \bibinfo{title}{Data integration for the assessment of population
		exposure to ambient air pollution for global burden of disease assessment}.
	\newblock \bibinfo{journal}{Environmental Science \& Technology}
	\bibinfo{volume}{52}, \bibinfo{pages}{9069--9078}.
	\newblock \DOIprefix\doi{10.1021/acs.est.8b02864}.
	\bibitem[{Snyder et~al.(2013)Snyder, Watkins, Solomon, Thoma, Williams, Hagler
		and Preuss}]{snyder2013changing}
	\bibinfo{author}{Snyder, E.G.}, \bibinfo{author}{Watkins, T.H.},
	\bibinfo{author}{Solomon, P.A.}, \bibinfo{author}{Thoma, E.D.},
	\bibinfo{author}{Williams, R.W.}, \bibinfo{author}{Hagler, G.S.},
	\bibinfo{author}{Preuss, P.W.}, \bibinfo{year}{2013}.
	\newblock \bibinfo{title}{The changing paradigm of air pollution monitoring}.
	\newblock \bibinfo{journal}{Environmental Science \& Technology}
	\bibinfo{volume}{47}, \bibinfo{pages}{11369--11377}.
	\bibitem[{Song et~al.(2021)Song, Han and Stettler}]{yoursource11}
	\bibinfo{author}{Song, J.}, \bibinfo{author}{Han, K.},
	\bibinfo{author}{Stettler, M.E.J.}, \bibinfo{year}{2021}.
	\newblock \bibinfo{title}{Deep-maps: Machine-learning-based mobile air
		pollution sensing}.
	\newblock \bibinfo{journal}{IEEE Internet of Things Journal}
	\bibinfo{volume}{8}, \bibinfo{pages}{7649--7660}.
	\newblock \DOIprefix\doi{10.1109/JIOT.2020.3041047}.
	\bibitem[{Wang et~al.(2014)Wang, Zhang, Niu and Liu}]{wang2014kalman}
	\bibinfo{author}{Wang, J.}, \bibinfo{author}{Zhang, L.}, \bibinfo{author}{Niu,
		F.}, \bibinfo{author}{Liu, Z.}, \bibinfo{year}{2014}.
	\newblock \bibinfo{title}{Effects of pm2.5 on health and economic loss:
		Evidence from beijing-tianjin-hebei region of china}.
	\newblock \bibinfo{journal}{Journal of Cleaner Production}
	\bibinfo{volume}{161}, \bibinfo{pages}{1508--1518}.
	\newblock \DOIprefix\doi{https://doi.org/10.1016/j.jclepro.2020.120605}.
	\bibitem[{Wang et~al.(2023)Wang, Lin, Chen, Zhang and
		Gao}]{wang2023forecasting}
	\bibinfo{author}{Wang, R.}, \bibinfo{author}{Lin, Y.}, \bibinfo{author}{Chen,
		J.}, \bibinfo{author}{Zhang, J.}, \bibinfo{author}{Gao, J.},
	\bibinfo{year}{2023}.
	\newblock \bibinfo{title}{Graphy: Physics-guided graph neural networks for
		fine-grained air quality forecasting}.
	\newblock \bibinfo{journal}{IEEE Transactions on Knowledge and Data
		Engineering} \DOIprefix\doi{https://doi.org/10.1145/3734869}.
	\bibitem[{Wang et~al.(2020)Wang, Li, Zhang et~al.}]{wang2020pm25gnn}
	\bibinfo{author}{Wang, S.}, \bibinfo{author}{Li, Y.}, \bibinfo{author}{Zhang,
		J.}, et~al., \bibinfo{year}{2020}.
	\newblock \bibinfo{title}{Pm2.5-gnn: A domain knowledge enhanced graph neural
		network for pm2.5 forecasting}, in: \bibinfo{booktitle}{SIGSPATIAL '20}, pp.
	\bibinfo{pages}{163--166}.
	\newblock \DOIprefix\doi{https://doi.org/10.1145/3397536.34222}.
	\bibitem[{Wen et~al.(2019)Wen, Liu, Yao, Peng, Li, Hu and Chi}]{wen2019deep}
	\bibinfo{author}{Wen, C.}, \bibinfo{author}{Liu, S.}, \bibinfo{author}{Yao,
		X.}, \bibinfo{author}{Peng, L.}, \bibinfo{author}{Li, H.},
	\bibinfo{author}{Hu, Y.}, \bibinfo{author}{Chi, T.}, \bibinfo{year}{2019}.
	\newblock \bibinfo{title}{A novel spatiotemporal convolutional long short-term
		neural network for air pollution prediction}.
	\newblock \bibinfo{journal}{Science of the Total Environment}
	\bibinfo{volume}{654}, \bibinfo{pages}{1091--1099}.
	\newblock \DOIprefix\doi{https://doi.org/10.1016/j.scitotenv.2018.11.086}.
	\bibitem[{Yi and Prybutok(2018)}]{yi2018review}
	\bibinfo{author}{Yi, J.}, \bibinfo{author}{Prybutok, V.}, \bibinfo{year}{2018}.
	\newblock \bibinfo{title}{A review of air quality forecasting techniques using
		machine learning models}.
	\newblock \bibinfo{journal}{International Journal of Environmental Research and
		Public Health} \bibinfo{volume}{15}, \bibinfo{pages}{2142}.
	\newblock \DOIprefix\doi{10.1016/j.jclepro.2021.129072}.
	\bibitem[{Zhang et~al.(2022)Zhang, Zheng, Xu, Chen and
		Wang}]{zhang2022attentioncnn}
	\bibinfo{author}{Zhang, W.}, \bibinfo{author}{Zheng, H.}, \bibinfo{author}{Xu,
		Y.}, \bibinfo{author}{Chen, Y.}, \bibinfo{author}{Wang, Y.},
	\bibinfo{year}{2022}.
	\newblock \bibinfo{title}{Attention-based convolutional neural network for
		fine-grained air quality prediction}.
	\newblock \bibinfo{journal}{Environmental Modelling \& Software}
	\bibinfo{volume}{147}, \bibinfo{pages}{105226}.
	\newblock \DOIprefix\doi{https://doi.org/10.1016/j.eswa.2024.125128}.
	\bibitem[{Zheng et~al.(2013)Zheng, Liu and Hsieh}]{zheng2013u}
	\bibinfo{author}{Zheng, Y.}, \bibinfo{author}{Liu, F.}, \bibinfo{author}{Hsieh,
		H.P.}, \bibinfo{year}{2013}.
	\newblock \bibinfo{title}{U-air: When urban air quality inference meets big
		data}, in: \bibinfo{booktitle}{Proceedings of the 19th ACM SIGKDD
		International Conference on Knowledge Discovery and Data Mining}, pp.
	\bibinfo{pages}{1436--1444}.
	\newblock \DOIprefix\doi{https://doi.org/10.1145/2487575.2488188}.
	\bibitem[{Zhou et~al.(2020)Zhou, Wang and Zhang}]{zhou2020forecasting}
	\bibinfo{author}{Zhou, G.}, \bibinfo{author}{Wang, P.}, \bibinfo{author}{Zhang,
		L.}, \bibinfo{year}{2020}.
	\newblock \bibinfo{title}{Forecasting {PM$_{2.5}$} concentrations using a
		spatiotemporal model with ensemble learning}.
	\newblock \bibinfo{journal}{Atmospheric Environment} \bibinfo{volume}{223},
	\bibinfo{pages}{117257}.
	\newblock \DOIprefix\doi{https://doi.org/10.1016/j.envint.2019.104909}.
	\bibitem[{Zhou et~al.(2024)Zhou, Liu, Chen and Yang}]{zhou2024deep}
	\bibinfo{author}{Zhou, Y.}, \bibinfo{author}{Liu, X.}, \bibinfo{author}{Chen,
		R.}, \bibinfo{author}{Yang, L.}, \bibinfo{year}{2024}.
	\newblock \bibinfo{title}{Deep learning-based air quality forecasting: A
		comprehensive review and future directions}.
	\newblock \bibinfo{journal}{Atmospheric Environment} \bibinfo{volume}{310},
	\bibinfo{pages}{119094}.
	\newblock \URLprefix
	\url{https://link.springer.com/article/10.1007/s40726-020-00159-z}.
	
\end{thebibliography}

\end{document}